%% file: main.tex
\documentclass{article}

\usepackage[preprint]{neurips_2024}

\usepackage{hyperref}
\hypersetup{colorlinks=true,linkcolor=red,urlcolor=teal,citecolor=blue,pdfstartview=FitH}

\usepackage[utf8]{inputenc} 
\usepackage[T1]{fontenc}    
\usepackage{url}            
\usepackage{booktabs}       
\usepackage{amsfonts}       
\usepackage{nicefrac}       
\usepackage{microtype}      
\usepackage{xcolor}         
\usepackage{graphicx}
\usepackage{array}
\usepackage{multirow}
\usepackage{caption}
\usepackage{subcaption}
\usepackage{wrapfig}
\usepackage{amsmath}

\usepackage{comment}

\def\inl{{\langle}}
\def\inr{{\rangle}}
\def\ienc{{\varphi}}
\def\tenc{{\psi}}
\def\beqa{\begin{eqnarray}}
\def\eeqa{\end{eqnarray}}
\def\beqann{\begin{eqnarray*}}
\def\eeqann{\end{eqnarray*}}
\def\loss{\ell}
\newcommand{\A}{\mathcal{A}}

\newcommand{\M}{\mathcal{M}}

\newcommand{\ntasks}{25 }

\newcommand{\itt}{I$\rightarrow$T}
\newcommand{\tti}{T$\rightarrow$I}

\title{Benchmarking Vision-Language Contrastive Methods for Medical Representation Learning}

\author{Shuvendu Roy\textsuperscript{1,2*}~~~~~~~Yasaman Parhizkar\textsuperscript{1,3}\thanks{Equal contribution}~~~~~~~Franklin Ogidi\textsuperscript{1}~~~~~~~\textbf{Vahid Reza Khazaie}\textsuperscript{1}\\
\textbf{Michael Colacci\textsuperscript{4}~~~~~~~Ali Etemad\textsuperscript{2}~~~~~~~Elham Dolatabadi\textsuperscript{1,3}~~~~~~~Arash Afkanpour\textsuperscript{1}}\thanks{Corresponding author. Email: arash.afkanpour@vectorinstitute.ai}\\
$^1$Vector Institute~~~~~~~
$^2$Queen's University, Canada\\
$^3$York University, Canada~~~~~~~
$^4$University of Toronto, Canada
}

\begin{document}

\maketitle

\vspace{-15pt}
\begin{abstract}
\vspace{-5pt}
We perform a comprehensive benchmarking of contrastive frameworks for learning multimodal representations in the medical domain. Through this study, we aim to answer the following research questions: (i) How transferable are general-domain representations to the medical domain? (ii) Is multimodal contrastive training sufficient, or does it benefit from unimodal training as well? (iii) What is the impact of feature granularity on the effectiveness of multimodal medical representation learning? To answer these questions, we investigate eight contrastive learning approaches under identical training setups, and train them on 2.8 million image-text pairs from four datasets, 
and evaluate them on 25 downstream tasks, including classification (zero-shot and linear probing), image-to-text and text-to-image retrieval, and visual question-answering. Our findings suggest a positive answer to the first question, a negative answer to the second question, and the benefit of learning fine-grained features. Finally, we make our code publicly available.
\end{abstract}

\section{Introduction}
\label{sec:intro}
Multimodal learning has become pivotal in recent AI advancements \citep{driess2023palm,girdhar2023imagebind,singh2020mmf}. Multimodal representation learning in particular, enhances representations by exploiting the interplay between modalities to improve overall performance. This has been demonstrated in both the general domain \citep{radford2021learning} as well as specialized areas such as the medical domain \citep{lin2023pmc,eslami2021does}.
The medical domain, in particular, has increasingly recognized the importance of multimodal representation learning due to the complex nature of medical decision-making, which involves processing data from multiple modalities (e.g., imaging, pathology, and text) when diagnosing and deciding on intervention plans \citep{acosta2022multimodal}. In this field, it is practically impossible to curate labeled datasets for all tasks, modalities, and outcomes necessary for training supervised models. Therefore, leveraging a large pool of unlabeled image-text data to build foundation models, which can be fine-tuned on numerous downstream tasks, marked a revolutionary advancement in the visual-language domain \citep{huang2023visual}.

A number of prior studies have prioritized dataset curation to advance contrastive learning in medical visual-language tasks, recognizing the importance of large, high-quality data in developing generalizable models \citep{ikezogwo2024quilt,lin2023pmc,eslami2021does}. While the size of pretraining data is critical for building a robust foundation model \citep{kaplan2020scaling}, the learning approach is equally important. 
In this context, given the availability of various possible routes for training foundation models specifically for the medical domain, we raise a number of key questions that have not been exhaustively and systematically studied in the literature. In particular, we raise three research questions (RQ) as follows.

\textbf{RQ1: How effective and transferable are general-domain representations to the medical domain?} By understanding the extent of knowledge transfer from general vision tasks to medical imaging tasks, we can capitalize on existing resources, potentially accelerating the development of specialized models. This understanding is critical for medical representation learning for two reasons: (a) it diminishes the necessity for large-scale medical datasets comparable to those in the general domain, and (b) it reduces the computational burden associated with training medical foundation models from scratch.

\textbf{RQ2. To train effective multimodal image-text models in the medical domain, is multimodal contrastive training sufficient, or should unimodal training be added for better performance?}
Given that it has been shown in the general vision domain that the addition of unimodal learning alongside multimodal learning can improve learned unimodal representations \citep{wei2024mmpareto,zhang2023multimodal}, we question whether this approach would benefit contrastive image-text training in the medical domain as well.

\textbf{RQ3. What is the impact of feature granularity on the effectiveness of multimodal representation learning in the medical domain?}
Learning both global (high-level) and local (low-level) representations is necessary for medical tasks \citep{lu2024visual,zhao2023clip}. For instance, detecting microcalcifications in mammograms involves identifying small, clustered specks that can indicate the early stages of breast cancer \citep{alsheh2019association}. Accordingly, we aim to identify the impact of different levels of feature granularity on contrastive representation learning in this context \citep{huang2021gloria,muller2022joint}. Although some prior studies have investigated the impact of granularity on a small scale, these studies have not been conducted under a unified framework with an identical training setup.

To address these research questions, we investigate eight main approaches to contrastive learning through extensive empirical analyses. First, we carefully curate an extensive set of datasets consisting of a total of 2.8 million image-text pairs. These datasets span three primary medical image modalities namely radiology (MRI, CT, ultrasound, and x-ray), histopathology, and endoscopy images. We then pretrain the contrastive methods on the paired curated samples, which is followed by evaluation on \ntasks downstream tasks, including classification (zero-shot and linear probing), image-to-text and text-to-image retrieval, and visual question-answering. In addition, we extend the evaluation to include two other modalities, including ophthalmology and dermatology.

Our key contributions are summarized below:

\hspace{2mm} $\bullet$ We perform a detailed and unified analysis of eight contrastive learning methods to study the transferability, the impact of unimodal training, and feature granularity
in the context of medical image-text representation learning. Our study involves training on 2.8 million pairs and evaluation on \ntasks downstream tasks across five imaging modalities, making it the first study of its kind and scale in the medical domain. 

\hspace{2mm} $\bullet$ Specifically, we find that (\textit{i}) representations learned from the general domain can be transferable to the medical domain, especially in a setting where the image encoder is partially frozen; (\textit{ii}) Unimodal representation learning does not help multimodal representation learning in medical downstream tasks; and (\textit{iii}) Fine-grained representation learning enhances multimodal medical representations.

\hspace{2mm} $\bullet$ To contribute to the area of medical image-text foundation models and to enable fast and accurate reproducibility, we open-source our codebase at \url{https://github.com/ShuvenduRoy/multimodal}

\section{Methods}

\subsection{Preliminaries}
To answer the questions raised above, we need to learn effective representations from paired image-text samples. Prior work has shown that contrastive learning is a suitable approach for learning representations from paired data \citep{radford2021learning, girdhar2023imagebind}. Let $\ienc$ denote an image encoder and $\tenc$ denote a text encoder that maps images and text to a common representation space, respectively. Given a batch of training samples $B = \{ (x_i, t_i)\}_{i=1}^N$, where $x_i$ and $t_i$ denote the $i^{\text{th}}$ image and text instances respectively, the InfoNCE loss \citep{oord2018representation} is optimized by minimizing the distance between the representations of an image and its corresponding text, $(\ienc(x_i), \tenc(t_i))$, while maximizing the distance between unrelated image-text representation pairs, $(\ienc(x_i), \tenc(t_j)), \hspace{1mm} i \neq j$:
\beqa
\label{eq:contrastive_loss}
\ell_{\text{con}}(x_i,t_i;B) = - \left( \log \frac{\exp(\inl \ienc(x_i), \tenc(t_i) \inr / \tau)}{\sum_{k=1}^N \exp (\inl \ienc(x_i), \tenc(\boldsymbol{t_k}) \inr / \tau)} +
 \log \frac{\exp(\inl \ienc(x_i), \tenc(t_i) \inr / \tau)}{\sum_{k=1}^N \exp (\inl \ienc(\boldsymbol{x_k}), \tenc(t_i) \inr / \tau)} \right),
\eeqa
where $\inl \cdot, \cdot \inr$ denotes similarity between two vectors (e.g. cosine similarity), and $\tau > 0$ is a temperature parameter. For simplicity of notation, we drop $B$ and denote the loss for $(x,t)$ by $\loss_\text{con}(x,t)$. Multimodal contrastive learning trains encoders $\ienc$ and $\tenc$ by minimizing Eq.~\ref{eq:contrastive_loss} over the pairs in $B$:
\beqa
\label{eq:multimodal_loss}
\loss_{\text{multimodal}} = \min_{\ienc, \tenc} \hspace{2mm} \mathbb{E}_B \Big[ \frac{1}{N} \sum_{i=1}^N  \ell_\text{con}(x_i,t_i)\Big].
\eeqa

\subsection{Explored solutions}

\begin{figure}[tb]
  \centering
  \includegraphics[width=1\textwidth]{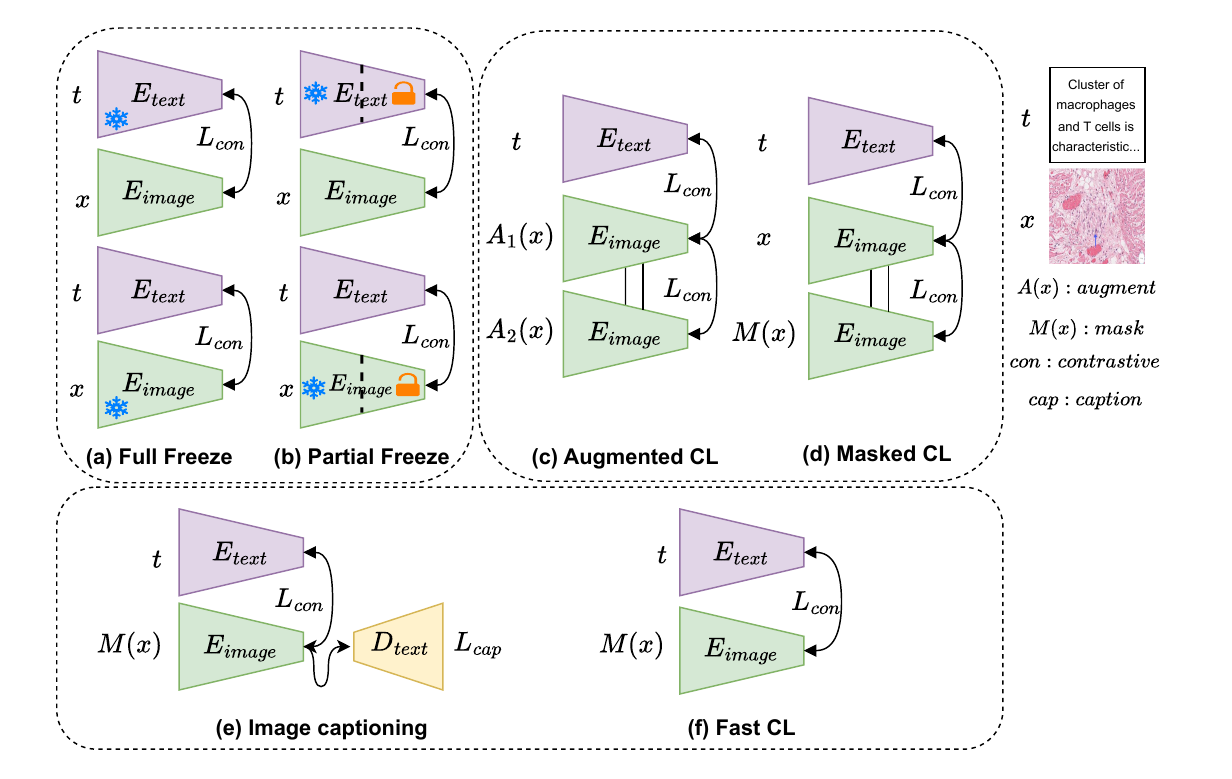}
  \caption{Illustration of eight contrastive learning approaches studied in the paper.}
  \label{fig:methods}
\end{figure}

\subsubsection{Transferability of general-domain representations}
To study the transferability and effectiveness of general-domain representations (RQ1), we consider encoders trained on a large-scale dataset of general image-text pairs. By a full or partial freeze of each of these encoders as shown in Figure~\ref{fig:methods} (a and b), we will study the effectiveness of representations of the corresponding modality for medical tasks. In particular, we consider the following cases:
\begin{enumerate}
    \item Freeze the first $\alpha$ portion of layers of the image encoder and the first $\beta$ portion of layers of the text encoder, but adapt the remaining layers. Here, \emph{adapt} refers to the unsupervised fine-tuning of a model (or part of it) via contrastive learning on medical image-text pairs.
    \item Fully freeze the image encoder, but adapt the text encoder ($\alpha = 1, \beta = 0$).
    \item Fully freeze the text encoder, but adapt the image encoder ($\alpha = 0, \beta = 1$).
\end{enumerate}
Partial fine-tuning while freezing early layers on a network allows for building high-level features on top of the early layers' features. In addition to leveraging general-domain features, this approach offers two additional benefits: (i) it reduces the number of learnable parameters, which improves robustness when training on a small amount of data, and (ii) it reduces the computational cost.

The extreme cases of a full encoder freeze (cases 2 and 3) imply a constant representation space for the corresponding modality. However, adaptation of the other modality's encoder allows for aligning its representations to that of the frozen encoder. A similar approach was taken in ImageBind \citep{girdhar2023imagebind}, where the text and image encoders were frozen, but encoders of four other modalities were tuned to align with the frozen encoders.

\subsubsection{Unimodal learning}
Vision-language contrastive learning maximizes alignment between image-text pairs. While this focuses on inter-modal alignment, a vast body of work in self-supervised learning has been devoted to unimodal representation learning. Leveraging unimodal representation learning in a multimodal framework has the potential to enhance representations by considering the relationship between examples within each modality \citep{wang2023one}. To explore this, we consider two unimodal contrastive learning approaches and combine each of them with multimodal contrastive loss during training.

For this purpose, we employ SimCLR \citep{chen2020simple} which is one of the pioneering techniques for visual self-supervised learning. In this technique, two views of an image are created by applying a series of random augmentations (e.g. crop-and-resize, color jitter, Gaussian blur, etc.) to the source image. Then, an encoder is learned by minimizing the contrastive loss (i.e., minimizing the distance between two views of an image in the representation space, and maximizing the distance between two views of different images) as illustrated in Figure~\ref{fig:methods} (c). Given a batch of data, $B$, and random augmentations, $\A_1$ and $\A_2$, SimCLR learns encoder $\ienc$ and projector $\xi$ (usually a multi-layer perceptron) by minimizing:
 \beqann
    \loss_{\text{unimodal}} = \min_{\ienc,\xi} \hspace{2mm} \mathbb{E}_{\A_1,\A_2,B} \Big[ \loss_\text{con} \Big( \xi(\ienc(\A_1(B))), \xi(\ienc(\A_2(B))) \Big) \Big].
\eeqann

Inspired by \citep{wang2023one}, we also perform contrastive learning between representations of an original image and its masked version, Figure~\ref{fig:methods} (d). Following \citep{dosovitskiy2020image}, we divide the image into non-overlapping patches and then use a masking operator, $\M$, to randomly mask a certain fraction of patches. The unimodal contrastive loss in this case is defined as,
\beqann
    \loss_{\text{unimodal}} = \min_{\ienc,\xi} \hspace{2mm} \mathbb{E}_{\M,B} \Big[ \loss_\text{con} \Big( \xi(\ienc(B)), \xi(\ienc(\M(B))) \Big) \Big].
\eeqann
Finally, in each case, the text and image encoders are trained by optimizing,
\beqann
    \lambda \loss_{\text{multimodal}} + (1-\lambda) \loss_{\text{unimodal}},
\eeqann
where, $\loss_{\text{multimodal}}$ is the image-text contrastive loss (Eq.~\ref{eq:multimodal_loss}), and $0 \leq \lambda \leq 1$ is a tradeoff parameter.

\subsubsection{Feature granularity in representation learning}
We study how changing the granularity of features affects multimodal representation learning.
To learn fine-grained features, we investigate CoCa \citep{yu2022coca}, which performs simultaneous image-text contrastive learning and image captioning as shown in Figure~\ref{fig:methods} (e). The addition of autoregressive caption generation via a text decoder conditioned on the image representation aims for detailed granularity \citep{yu2022coca}. The loss is defined as $\lambda \loss_\text{multimodal} + (1-
\lambda) \loss_{\text{cap}}$, where for an image-text pair $(x, t)$, the captioning loss, $\loss_{\text{cap}}$, is defined as,
\beqann
    \loss_{\text{cap}}(x,t) = -\sum_{\nu} \log \delta(t_{\nu}|t_{< \nu}, \ienc(x)).
\eeqann
Here, $\lambda$ is a trade-off parameter, $\delta$ is a text decoder, $t_{\nu}$ denotes the $\nu$-th token of $t$,  $t_{< \nu}$ denotes all tokens up to the $\nu$-th token in $t$, and $\ienc(x)$ denote the representation vector of $x$.

Going in the other direction, to study the effect of learning high-level (coarse-grained) features in multimodal contrastive learning, we follow FLIP \citep{li2023scaling}. This method masks a large portion of image patches and applies contrastive loss between the visible patches and text, Figure~\ref{fig:methods} (f). In addition to enforcing the vision encoder to learn higher-level features, masking 50-75\% of patches also reduces the computation cost by 2-4$\times$.

\section{Experiments}
In this section, we describe the experimental setup used to pretrain, fine-tune, and evaluate contrastive vision-language models. Then, we will present our results and findings.

\subsection{Setup}
\begin{figure}
    \centering
    \includegraphics[width=0.9\textwidth]{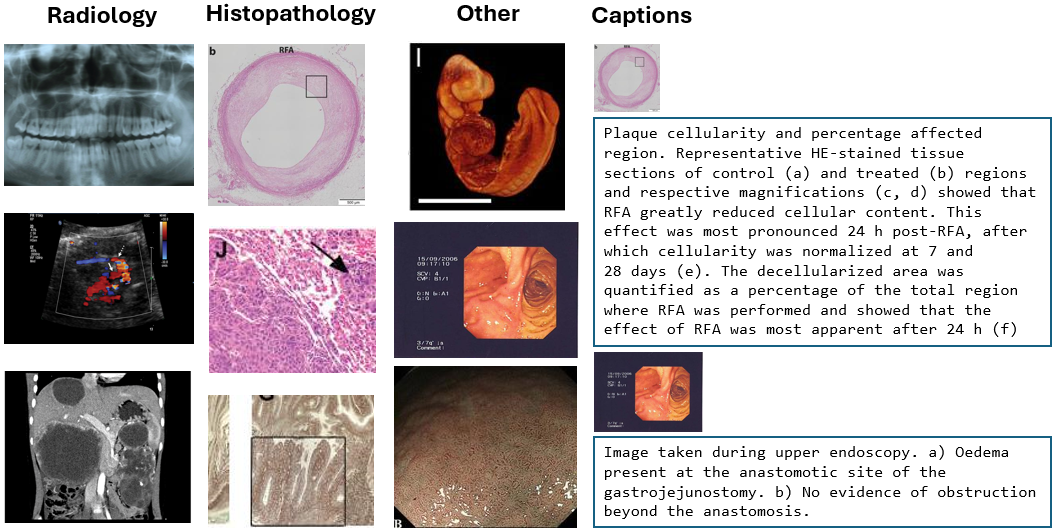}
    \caption{Samples from the datasets used in this study.}
    \label{fig:samples}
\end{figure}

\textbf{Pretraining.} We train \textit{eight} contrastive vision-language models, categorized along three key dimensions: transferability (RQ1), unimodal learning (RQ2), and fine-grained representation learning (RQ3). Table~\ref{tab:model_descriptions} presents the benchmarked methods and their descriptions. Here, \textbf{Fast CL} follows the prescription of contrastive learning between a masked image and the corresponding text, proposed by FLIP \citep{yao2021filip}. The training data consists of 2.8 million image-text pairs encompassing radiology (1.4M), histopathology (1.2M), and general medical images (e.g., endoscopy). Please see Figure ~\ref{fig:samples} for a few samples. These pairs are sourced from four medical datasets including PMC-OA \citep{lin2023pmc}, Quilt-1M \citep{ikezogwo2024quilt}, MIMIC-CXR \citep{johnson2019mimic}, and ROCO \citep{pelka2018radiology}.
These datasets contain a broad spectrum of textual content such as scholarly articles, clinical reports, and social media posts. We use the training splits of Quilt-1M, MIMIC-CXR, and ROCO for pretraining, while the test splits are later used for retrieval in downstream evaluation. The PMC-OA dataset is entirely used for pretraining. Please refer to Section \ref{sec:appendix-datasets} for more details on the pretraining data. 

\begin{table}[tb]
    \caption{Different contrastive learning (CL) methods and their description.}
    \label{tab:model_descriptions}
    \centering
    \small
    \begin{tabular}{l|l|l}
        \midrule
        \textbf{RQ} & \textbf{Method name} & \textbf{Description}\\ \midrule
        \multirow{4}{*}{RQ1} & \textbf{Image Partial Freeze} & Freeze the first 50\% of the image encoder  \\
        & \textbf{Text Partial Freeze} & Freeze the first 50\% of the text encoder  \\
        & \textbf{Image Full Freeze} & Fully freeze the image encoder  \\
        & \textbf{Text Full Freeze} & Fully freeze the text encoder  \\
        \midrule
        \multirow{2}{*}{RQ2} & \textbf{Augmented CL} & Combined unimodal (SimCLR \citep{chen2020simple}) and multimodal CL \\
        & \textbf{Masked CL} & Combined unimodal masked and multimodal CL \\
        \midrule
        \multirow{2}{*}{RQ3}  & \textbf{Image Captioning} & Combined CL and captioning \citep{yu2022coca} \\
        & \textbf{Fast CL} & CL between text and the masked image \citep{li2023scaling} \\
        \bottomrule
    \end{tabular}
\end{table}

\textbf{Evaluation.} We perform an extensive evaluation of all models on a total of \ntasks retrieval, classification, and visual question-answering (VQA) tasks. Within our evaluation, we also explore models' generalization performances on two modalities including dermatology and ophthalmology that are not 
present in pretraining data. Table \ref{tab:eval-data} presents the datasets used for all evaluations. The image and text retrieval tasks encompass both image-to-text (\itt) and text-to-image (\tti) retrieval. To evaluate the (\itt) performance, we utilize test sets from the ROCO, Quilt, and MIMIC-CXR datasets. For each case, we report Recall@200 results. The image classification tasks include both zero-shot classification and linear probing across 22 tasks, including histopathology (9 tasks), radiology (8 tasks), dermatology (3 tasks), and ophthalmology (2 tasks). Visual question-answering is performed using the Mixture of Enhanced Visual Features (MEVF) method \citep{do2021multiple} without using a decoder. We use our trained vision and text encoders to encode the image and question, respectively. This task contain open- and close-ended questions and is mapped as a classification problem and performed on PathVQA \citep{he2020pathvqa}, VQARAD \citep{lau2018dataset} datasets.

\begin{table}[t]
\small
  \caption{Evaluation datasets used in this study.}
  \label{tab:eval-data}
  \resizebox{\textwidth}{!}{
  \centering
  \begin{tabular}{m{2cm} m{2cm} m{3cm} m{3.5cm} c}
    \toprule
    \textbf{Task} & \textbf{Setup} & \textbf{Dataset} & \textbf{Modality} & \textbf{Nb. Samples}  \\
    \midrule
    \multirow{4}{2cm}{Retrieval} & \multirow{4}{2cm}{\itt ~\& \tti} & Quilt-1M & Histopathology & 13,559 \\
    &  & MIMIC-IV-CXR & Chest X-ray & 3,269  \\
    &  & ROCO & Chest X-ray & 8,176  \\
    &  & DeepEyeNet & Retina & 3,140 \\
    \midrule
    \multirow{22}{2cm}{ Zero-shot classification \& Linear probing} 
    & 10 classes* & VinDr-Mammo & Mammography & 4,095  \\
    & 6 classes & PAD-UFES-20 & Dermatology & 460  \\
    & 7 classes & SkinCancer & Dermatology & 2,003  \\
    & 2 classes & PatchCamelyon (PCam) & Histopathology & 32,768  \\
    & 8 classes & NCT-CRC-HE-100K & Histopathology & 6,333  \\
    & 3 classes & LC25000Lung & Histopathology & 3,000  \\
    & 2 classes & LC25000Colon & Histopathology & 2,000 \\
    & 4 classes & BACH & Histopathology & 100 \\
    & 4 classes & SICAPv2 & Histopathology & 2,122  \\

    & 14 classes* & ChestMNIST+ & Chest X-ray & 112,120  \\
    & 9 classes & PathMNIST+ & Colon Pathology & 107,180  \\
    & 7 classes & DermaMNIST+ & Dermatoscope & 10,015  \\
    & 4 classes & OctMNIST+ & Retinal OCT & 109,309  \\
    & 2 classes  & PneumoniaMNIST+ & Chest X-Ray & 5,856  \\
    & 5 classes & RetinaMNIST+ & Fundus Camera & 1,600 \\
    & 2 classes  & BreastMNIST+ & Breast Ultrasound & 780  \\
    & 8 classes & BloodMNIST+ & Blood Cell Microscope & 17,092  \\
    & 8 classes & TissueMNIST+ & Kidney Cortex Microscope & 236,386  \\
    & 11 classes & OrganAMNIST+ & Abdominal CT & 58,830  \\
    & 11 classes & OrganCMNIST+ & Abdominal CT & 23,583  \\
    & 11 classes & OrganSMNIST+ & Abdominal CT & 25,211 \\
    \midrule
    \multirow{2}{2cm}{VQA} 
    & \multirow{2}{2cm}{-} & VQA-RAD & Radiology & 3,515 \\
    & & PathVQA & Histopathology & 32,799 \\
    \bottomrule
    \multicolumn{2}{l}{*Denotes multi-label classification.} 
  \end{tabular}
  }
\vspace{-10pt}
\end{table}

\begin{table}[tb]
    \caption{Vision encoder selection results. For each case we measure validation loss on PMC-OA validation set, retrieval Recall@200 on four datasets, and VQA accuracy across all answer types on two datasets. Best values are highlighted in bold.}
    \label{tab:encoder_combination}
    \resizebox{\textwidth}{!}{
    \centering
    \small
    \begin{tabular}{@{}c|c|cc|cc|cc|cc||cc@{}}
\toprule
\multirow{2}{*}{\textbf{Vision encoders}} & \multirow{2}{*}{\textbf{Val. loss}} & \multicolumn{2}{c}{\textbf{ROCO}} & \multicolumn{2}{c}{\textbf{Quilt}} & \multicolumn{2}{c}{\textbf{MIMIC-CXR}} & \multicolumn{2}{c||}{\textbf{DeepEyeNet}} & \textbf{PathVQA} & \textbf{VQARAD} \\ 
 &  & \multicolumn{1}{l}{\itt} & \tti & \multicolumn{1}{l}{\itt} & \tti & \multicolumn{1}{l}{\itt} & \tti & \multicolumn{1}{l}{\itt} & \tti & overall & overall \\ \midrule
RN-50 & 1.31 & 0.68 & 0.72 & 0.32 & 0.35 & 0.25 & 0.40 & 0.09 & 0.10 & 46.44 & 62.53 \\
ViT-B/16 & \textbf{1.15} & \textbf{0.78} & \textbf{0.77} & \textbf{0.44} & \textbf{0.46} & \textbf{0.27} & \textbf{0.44} & 0.11 & \textbf{0.11} & \textbf{46.75} & 61.64 \\
ViT-B/32 & 1.94 & 0.72 & 0.71 & 0.42 & 0.43 & 0.33 & 0.43 & \textbf{0.12} & 0.10 & 46.33 & \textbf{62.97} \\ \bottomrule
    \end{tabular}
    }
\end{table}

\subsection{Model selection}
Due to the extensive nature of this study, it is challenging to perform model selection for each contrastive learning variant. Therefore, we first perform hyperparameter tuning and model selection with vanilla contrastive learning. Hyperparameter tuning was mainly performed as a grid search on batch size and learning rate. More details are available in Section \ref{sec:appendix-hparam-config}.
We evaluate three image encoders, ResNet-50 \citep{he2016deep}, ViT-B/16, and ViT-B/32 transformers \citep{dosovitskiy2020image}. For the text encoder we used the default GPT/77 encoder used by CLIP \citep{radford2021learning}. 
We pretrain each candidate on our pretraining data (2.8M pairs) and then evaluate its performance by measuring validation loss, retrieval Recall@200, and VQA accuracy. The validation loss is measured on the validation set of the PMC-OA dataset. The retrieval performance is measured for both image-to-text and text-to-image cases on ROCO, Quilt, MIMIC-CXR, and DeepEyeNet datasets. VQA accuracy is measured on PathVQA and VQARAD datasets. For VQA tasks we use a context length of 12 instead of 77 for the text encoder, similar to \citep{do2021multiple}. Table~\ref{tab:encoder_combination} presents the results for these encoders.
Findings from this experiment suggest that ViT-B/16 is the best-performing encoder compared to other encoders in our study. Specifically, it achieves the best validation loss and outperforms the other encoders on 6 out of 8 retrieval and 1 out of 2 VQA tasks. Therefore, we consider ViT-B/16 as our default vision encoder and consider the performance of this model as the baseline for other experiments throughout the paper.

\subsection{Findings}

In general, we find a positive response to our first research question, and a negative response to the second research question. Finally, findings on the third question suggest that learning fine-grained representations can be beneficial over course-grained representation for medical domain.  Further details are provided below.

\begin{figure}[t]
    \centering
    \includegraphics[width=0.83\linewidth]{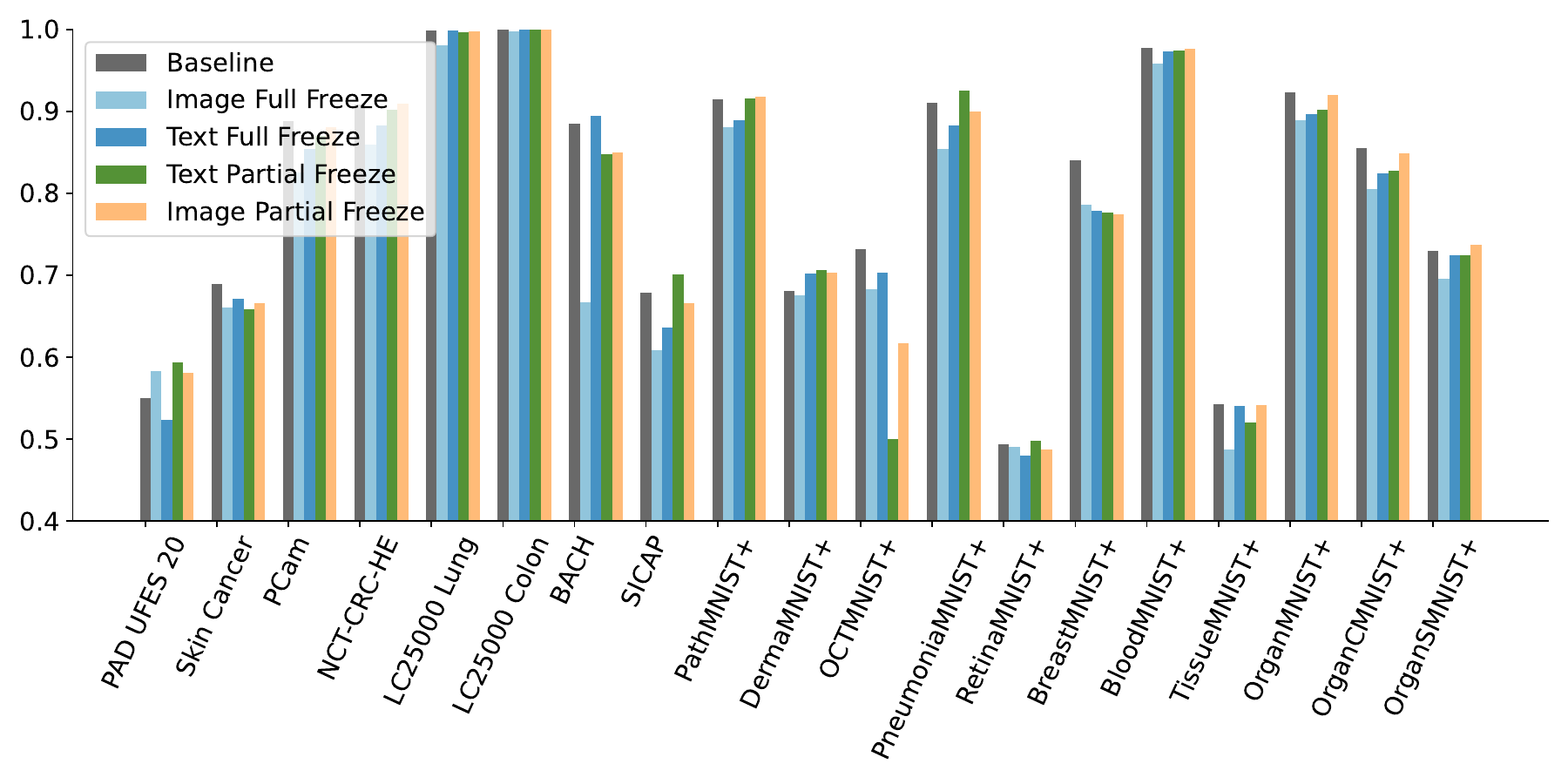}
    \caption{F1 score for linear probing on the study of transferability. Here, Image Partial Freeze outperforms the baseline on 6 datasets, while Text Partial Freeze, Text Full Freeze and Image Full Freeze perform better than the baseline in 5, 3, and 1 datasets, respectively. }
    \label{fig:f1_rq1}
\end{figure}

\textbf{Representations learned in the general domain \textit{can} be transferred to the medical domain.}
As summarized in Table \ref{tab:model_descriptions}, we investigate four contrastive learning approaches to examine the transferability of representations from the general domain to the medical domain. The retrieval and VQA results are presented in Table \ref{tab:ret_results_1}, while the classification results are shown in Figure \ref{fig:f1_rq1}. 
We observe that in 12 tasks (6 in retrieval and 6 in classification), partially freezing the image encoder outperforms the baseline. 
These findings suggest that transferability from the general domain is a viable route for representation learning in the medical domain. In particular, primitive features learned in the early layers of a vision transformer on a large, general domain dataset do not require much adaptation by further pretraining on medical data. Accordingly, a partial freeze of the image encoder trained on general domain data could lead to performance improvement in the medical domain. 
The same phenomenon, however, is not observed when partially freezing the text encoder. 
Only 6 fine-tuning tasks (5 in classification and 1 in VQA) benefit from partial freezing of the text encoders. 
Finally, fully freezing either of the encoders hurts performance. This is not unexpected, as representations learned for general-domain data may not be fully transferrable to the medical domain. Yet, for full encoder freezing, 5 tasks (3 in classification and 2 in retrieval) benefit from this approach for text encoders, while 1 classification task showed improvement with full freezing of image encoders.

\begin{table}[tb]
\centering
\small
\setlength\tabcolsep{4pt}
\caption{Retrieval (Recall@200) and VQA performance of various methods examined in RQ1 for transferability of learned representation from the general domain.}
\label{tab:ret_results_1}
    \small
    \begin{tabular}{@{}c|cc|cc|cc|cc||cc@{}}
\toprule
\multirow{2}{*}{\textbf{Models}} & \multicolumn{2}{c|}{\textbf{\footnotesize{ROCO}}} & \multicolumn{2}{c|}{\textbf{\footnotesize{Quilt}}} & \multicolumn{2}{c|}{\textbf{\footnotesize{MIMIC-CXR}}} & \multicolumn{2}{c||}{\textbf{\footnotesize{DeepEyeNet}}} & \multicolumn{1}{l}{\textbf{\footnotesize{PathVQA}}} & \multicolumn{1}{l}{\textbf{\footnotesize{VQARAD}}} \\
 & \itt & \tti & \itt & \tti & \itt & \tti & \itt & \tti & overall & overall \\ \midrule
Baseline & 0.78 & 0.77 & 0.44 & 0.46 & 0.27 & 0.44 & 0.11 & 0.11 & \textbf{46.75} & 61.24 \\
Image Full Freeze & 0.52 & 0.61 & 0.22 & 0.28 & 0.17 & 0.24 & 0.07 & 0.10 & 46.28 & 59.27 \\
Text Full Freeze & 0.64 & 0.65 & 0.28 & 0.28 & 0.24 & 0.25 & \textbf{0.12} & \textbf{0.12} & 46.69 & 59.87 \\
Image Partial Freeze & \textbf{0.82} & \textbf{0.82} & \textbf{0.48} & \textbf{0.51} & \textbf{0.43} & \textbf{0.55} & 0.07 & 0.11 & 46.06 & 61.42 \\
Text Partial Freeze & 0.73 & 0.72 & 0.40 & 0.41 & 0.32 & 0.43 & 0.09 & 0.11 & 46.31 & \textbf{63.19} \\ \bottomrule
\end{tabular}
\end{table}

\begin{figure}[t]
    \centering
    \includegraphics[width=0.83\linewidth]{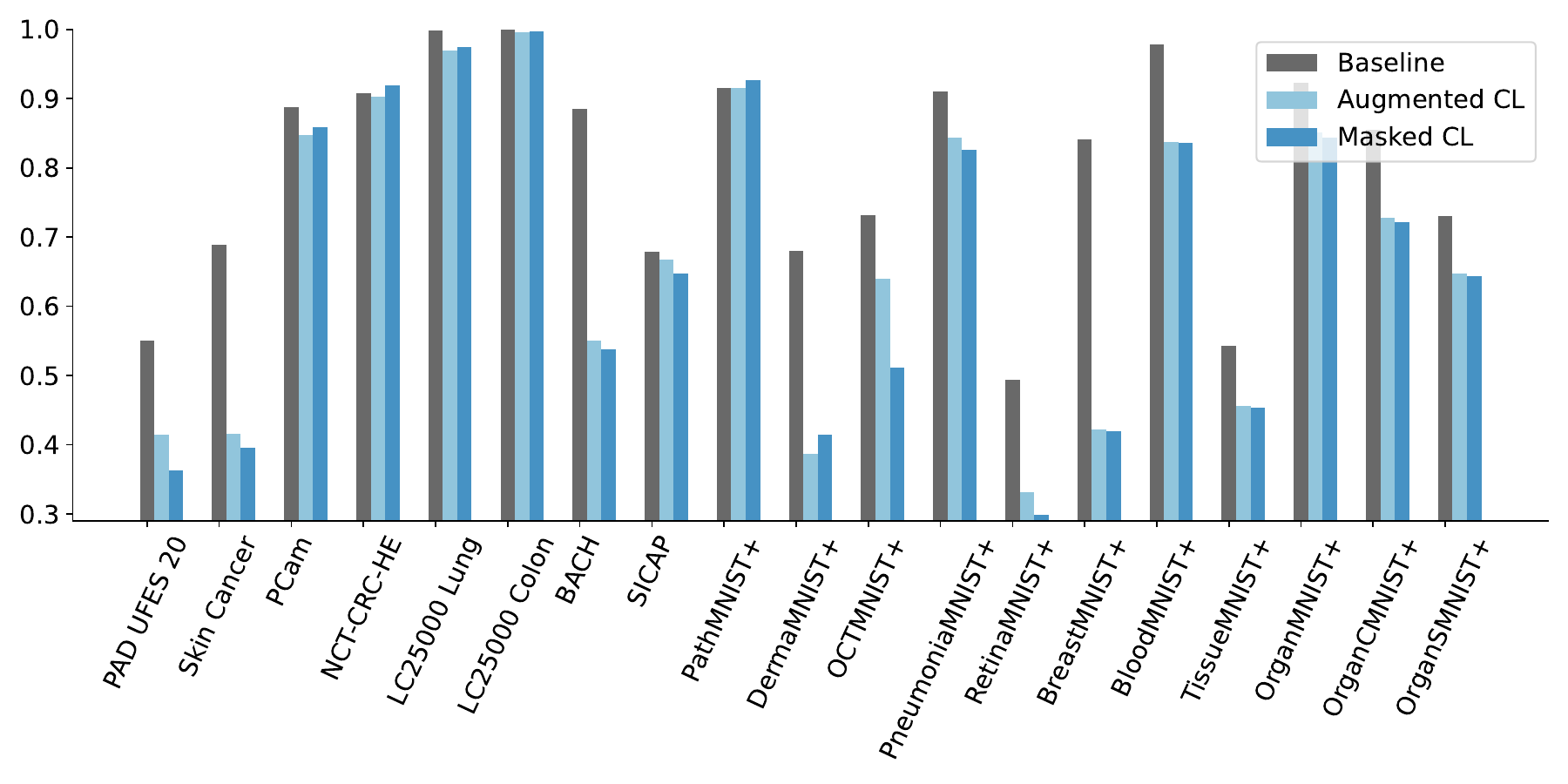}
    \caption{F1 score for linear probing on the study of unimodal learning. Here, Masked CL performs better than the baseline on two datasets. }
    \label{fig:f1_rq2}
\end{figure}

\textbf{Unimodal representation learning \textit{may not} enhance multimodal learning.}
The results of our experimentation on two contrastive learning approaches enabling joint unimodal and multimodal representation learning are presented in Table \ref{tab:rq2} (retrieval and VQA) and Figure \ref{fig:f1_rq2} (classification). While neither of the variants in this study perform well across all tasks, masked CL does improve performance in the 2 VQA tasks, 2 classification tasks, and one retrieval task (\itt). A few explanations are possible. For instance, a different trade-off in the loss combination could result in a better performance. It is also possible that optimizing the unimodal and multimodal loss functions requires very different feature sets, making the resulting representation space ineffective for our downstream tasks. Further investigation is required to better understand these results.

\begin{table}[t]
\caption{Retrieval (Recall@200) and VQA performance of various methods examined in RQ2 for integrating unimodal representation learning with multimodal learning. }
\small
\centering
\label{tab:rq2}
    \resizebox{\textwidth}{!}{
    \begin{tabular}{@{}c|cc|cc|cc|cc||cc@{}}
    \toprule
    \multirow{2}{*}{\textbf{Models}} & \multicolumn{2}{c|}{\textbf{ROCO}} & \multicolumn{2}{c|}{\textbf{Quilt}} & \multicolumn{2}{c|}{\textbf{MIMIC-CXR}} & \multicolumn{2}{c||}{\textbf{DeepEyeNet}} & \textbf{PathVQA} & \textbf{VQARAD} \\
 & \itt & \tti & \itt & \tti & \itt & \tti & \itt & \tti & overall & overall \\ \midrule
Baseline & \textbf{0.78} & \textbf{0.77} & \textbf{0.44} & \textbf{0.46} & \textbf{0.27} & \textbf{0.44} & 0.11 & \textbf{0.11} & 46.75 & 61.64 \\
Augmented CL & 0.56 & 0.55 & 0.32 & 0.34 & 0.21 & 0.27 & 0.11 & 0.08 & 46.25 & 63.24 \\
Masked CL & 0.56 & 0.56 & 0.33 & 0.35 & 0.21 & 0.28 & \textbf{0.13} & 0.09 & \textbf{46.86} & \textbf{64.52} \\ \bottomrule
    \end{tabular}
    }
\end{table} 
\begin{table}
\caption{Retrieval performance (Recall@200) of various methods examined in RQ3 for feature granularity.}
\centering
\small
\label{tab:rq3}
    \resizebox{\textwidth}{!}{
    \begin{tabular}{@{}c|cc|cc|cc|cc||cc@{}}
    \toprule
    \multirow{2}{*}{\textbf{models}} & \multicolumn{2}{c|}{\textbf{ROCO}} & \multicolumn{2}{c|}{\textbf{Quilt}} & \multicolumn{2}{c|}{\textbf{MIMIC-CXR}} & \multicolumn{2}{c||}{\textbf{DeepEyeNet}} & \textbf{PathVQA} & \textbf{VQARAD} \\
 & \itt & \tti & \itt & \tti & \itt & \tti & \itt & \tti & overall & overall \\ \midrule
Baseline & \textbf{0.78} & \textbf{0.77} & \textbf{0.44} & \textbf{0.46} & 0.27 & 0.44 & \textbf{0.11} & 0.11 & 46.75 & 61.64 \\
Fast CL (50\%) & 0.77 & 0.76 & 0.43 & 0.45 & \textbf{0.34} & \textbf{0.5} & 0.08 & 0.09 & 46.25 & 61.64 \\
Image captioning & 0.74 & 0.74 & \textbf{0.44} & \textbf{0.46} & \textbf{0.34} & 0.47 & 0.09 & \textbf{0.12} & \textbf{47.46} & \textbf{62.53} \\       \bottomrule
    \end{tabular}
    }
\end{table} 

\begin{figure}
    \centering
    \includegraphics[width=0.83\linewidth]{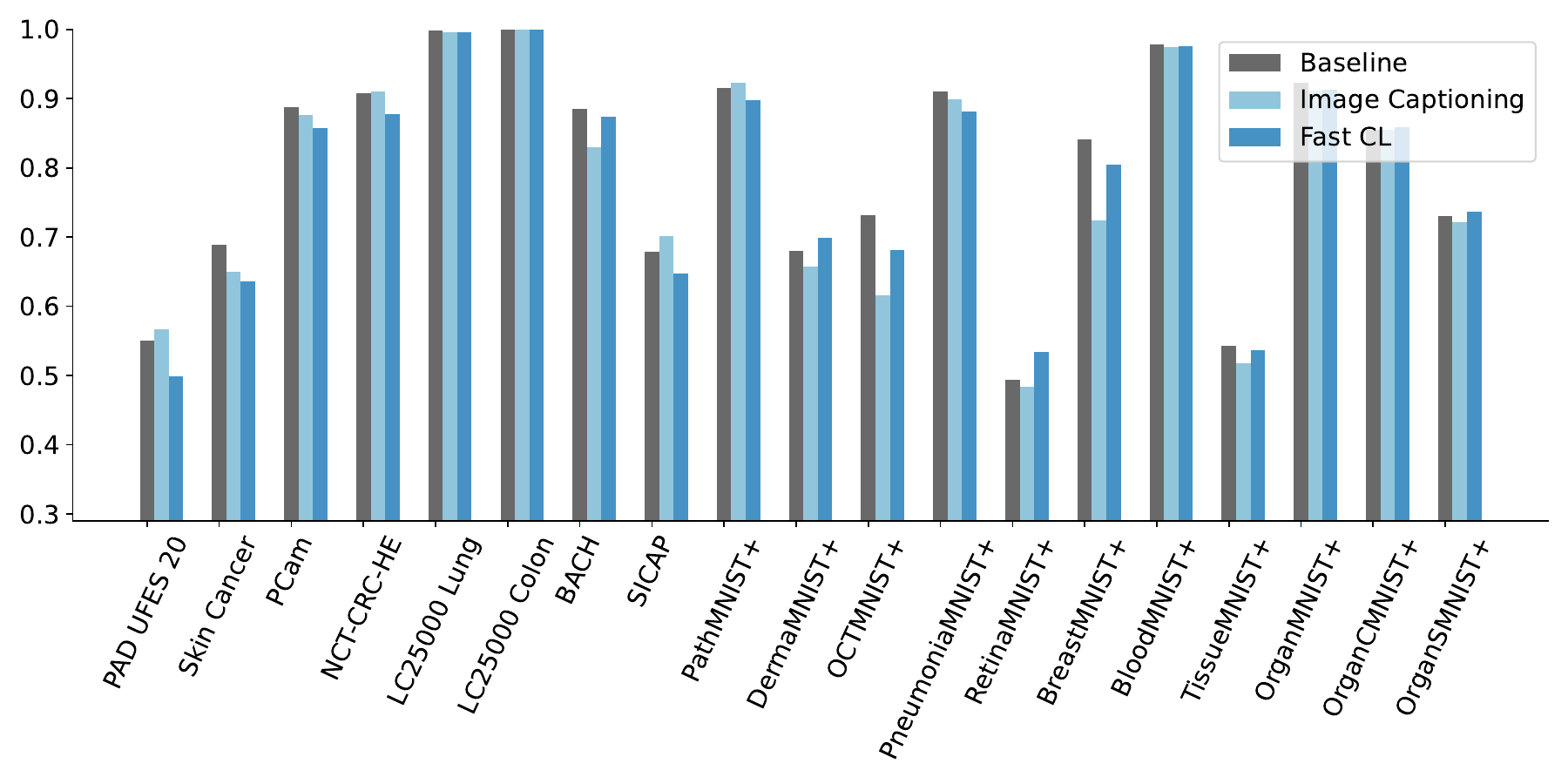}
    \caption{F1 score for linear probing on the study of feature granularity. Here, Image Captioning and Fast CL outperform the baseline on 4 datasets each. }
    \label{fig:f1_rq3}
\end{figure}

\textbf{Fine-grained representation learning \textit{can enhance} multimodal medical representations.}
In this study, Fast CL and Image Captioning are the two contrastive learning methods aimed at encouraging learning of the coarse-grained and fine-grained representations, respectively. As shown in Table \ref{tab:rq3}, fine-grained learning improved performance in 4 retrieval tasks and 2 VQA tasks. Additionally, Figure \ref{fig:f1_rq3} demonstrates that learning fine-grained features enhanced performance in 4 classification tasks. On the other hand, Fast CL outperforms the baseline in 2 retrieval tasks and 4 classification tasks. These results suggest that learning fine-grained features could be useful in the medical domain. This observation is supported by the fact that local details are often important in medical diagnosis tasks and aligned with the findings of previous works showing the importance of learning fine-grained features \citep{lu2024visual}.

\textbf{Other observations.} 
Based on a holistic view of the results obtained from the experiments, we observe that in the medical domain, for the task of retrieval, Image Partial Freezing yields the overall highest results based on Tables \ref{tab:ret_results_1}, \ref{tab:rq2} and \ref{tab:rq3}. For both zero-shot learning and linear probing, the standard CLIP baseline shows the overall best results according to Figures \ref{fig:f1_rq1} through \ref{fig:f1_rq3}. Finally, for VQA, according to Tables \ref{tab:rq2} and \ref{tab:rq3}, Masked CL and Image Captioning can yield the best performances.

\section{Related Work}
CLIP~\citep{radford2021learning}  stands as a pioneering work in multimodal visual and text representation learning, leveraging contrastive cross-entropy. Its application has extended across diverse domains, including the medical field \citep{boecking2022making, tiu2022expert, seibold2022breaking,liu2024etp,zhang2023text,liu2023imitate,kim2023concept}. One notable follow-up study, LiT~\citep{zhai2022lit}, proposed contrastive tuning with a frozen image encoder, demonstrating an approach to enhance pretraining efficiency without extensive datasets. Another study, FLIP~\citep{li2023scaling} involves randomly masking out a substantial portion of image patches during training. This simple technique encourages the model to learn higher-level representations, leading to improved performance. Another noteworthy technique is CoCa~ \citep{yu2022coca} proposed an efficient learning approach by jointly training with contrastive and captioning (generative) losses. In the medical domain, a number of studies \citep{huang2021gloria,muller2022joint,lu2024visual} showed that incorporating fine-grained contrastive learning improves performance.

\section{Conclusion}
We performed an extensive study of eight contrastive learning approaches for multimodal representation learning in the medical domain. This includes evaluating these variants on 19 classification tasks, 4 retrieval tasks, and 2 VQA tasks. Our results demonstrate that partial freezing of early layers of vision transformers and fine-tuning the remaining layers via contrastive learning could improve performance in medical downstream tasks. Additionally, incorporating techniques in contrastive learning that lead to fine-grained visual features could potentially improve performance. Given the relatively small volume of multimodal medical data, future work should study other methods to improve the performance of contrastive learning in the medical domain.

\textbf{Limitations.}
There are many possibilities for improving upon vanilla contrastive learning beyond what we studied in this paper. Many alternatives can be considered for augmenting multimodal learning with unimodal representation learning. Also, there are many other ways to encourage learning features of different granularities beyond what we studied here. Additionally, there are other important data modalities in healthcare, such as Electronic Health Records, time series, and tabular data, which could be covered in future work. Broader impact is discussed in Section~\ref{sec:limitations} in the Appendix.

\bibliographystyle{apalike}
\bibliography{main}

\clearpage
\appendix
\input{appendix}

\end{document}

%% file: appendix.tex
\section{Appendix}

\subsection{Related Work}\label{RW}

\textbf{Vision language contrastive learning}.
Learning multimodal visual and text representations without human supervision using contrastive cross-entropy loss is a straightforward yet powerful pretraining paradigm that has gained significant interest in self-supervised tasks. The pioneering work of CLIP \citep{radford2021learning} builds a task-agnostic model by leveraging this method, which performs competitively with task-specific supervised models \citep{zhang2024vision}. ALIGN \citep{jia2021scaling} further scales up CLIP by leveraging a noisy dataset containing over one billion image-alt-text pairs.
Given the data and compute-intensive nature of these Vision-Language Models (VLMs), several studies have explored more data and compute-efficient VLM pretraining methods using fewer image-text pairs. DeCLIP \citep{li2021supervision} introduces nearest-neighbor supervision to utilize information from similar pairs, enabling effective pretraining on limited data. LiT \citep{zhai2022lit} proposed contrastive tuning with a frozen image encoder, demonstrating another approach to enhance pretraining efficiency without extensive datasets.
FLIP proposes a simple method of randomly masking out a large portion of image patches during training, which improves performance by encouraging the model to learn higher-level representations \citep{li2023scaling}.
Another line of research has proposed combining dual-encoder contrastive pretraining with unimodal pretraining \citep{li2021align,yu2022coca,singh2022flava}. This approach allows models to inherit the strengths of both methods, resulting in strong performance on vision-language benchmarks.  For instance, ALBEF \citep{li2021align} integrates contrastive loss with masked language modeling using a dual-encoder design. CoCa \citep{yu2022coca} presents a simpler and more efficient approach by jointly training with contrastive and captioning (generative) losses. Another line of follow-up studies has focused on capturing finer-level information, such as the relationship between visual objects and textual words, by performing image-text contrastive learning across various semantic levels \citep{zhan2021product1m,li2020unimo,kim2021vilt}. For example, FILIP \citep{yao2021filip} introduces fine-grained semantic alignment (image patches and text tokens) through a novel cross-modal late interaction mechanism in contrastive learning, enabling the model to learn detailed vision-language correspondence.

\textbf{Contrastive learning in the medical domain}. 
The development and extension of contrastive vision language representations, particularly through models like CLIP \citep{zhao2023clip}, has garnered significant interest in the medical domain. These models have demonstrated promising results and explored transferability of original CLIP in their application to chest X-ray \citep{boecking2022making, tiu2022expert, seibold2022breaking}, ECG data \citep{liu2024etp}, histological analysis of stomach tissue \citep{zhang2023text}, lung CT scans \citep{liu2023imitate}, dermatology \citep{kim2023concept}, and eye fundus images \citep{baliah2023exploring}.
To satisfy the needs of medical CLIP-style (or foundational models) and resolve data scarcity, some studies have focused on generating novel image-text pairs across multiple domains \citep{lin2023pmc,eslami2021does,zhang2023biomedclip} and histopathology \citep{ikezogwo2024quilt,huang2023visual}. Due to notable differences between general web images and medical domain images, where small abnormalities can significantly influence diagnostic results, incorporating fine-grained contrastive learning has been shown to improve performance \citep{chaitanya2020contrastive}. Similarly, in diagnostic reports consisting of multiple sentences that describe image findings in specific regions, the inclusion of local vision features has been demonstrated to be important for accurate interpretation \citep{pang2023survey}. Studies such as GLoRIA \citep{huang2021gloria} and LoVT \citep{muller2022joint} have explored local-level cross-modal contrastive learning between both text-to-image and image-to-text local features. Additionally, CONCH \citep{lu2024visual} incorporated (generative) loss for fine-grained region-level features, thereby benefiting various tasks such as visual recognition, crossmodal alignment, image captioning, and multimodal understanding.

\textbf{Benchmarking in vision-language contrastive learning}.
Benchmarking is recognized as an essential tool for continuous improvement of quality \citep{dattakumar2003review}. In the deep learning era, characterized by rapid advancements and the proliferation of novel methods, the role of benchmarking has become increasingly pronounced \citep{russakovsky2015imagenet}. It has evolved to provide a fair and consistent basis for evaluating different models, facilitating comparison, and driving innovation \citep{chen2024benchmarking}. In the realm of vision-language contrastive learning, surprisingly, there have been only a few comprehensive benchmark studies. In a recent study, \citet{cui2022democratizing} benchmarked contrastive learning across the three dimensions of data, supervision, and model architecture. They discovered that the quality of the data and the application of appropriate supervision significantly enhance the performance of contrastive learning. \citet{tu2024closer} explored the safety objectives of 83 CLIP models in terms of resilience to variations in visual factors, calibrated uncertainty estimations, and the ability to detect anomalous inputs. The differences between these models were not in their architecture but in their training methods, usage scenarios (such as few-shot learning and fine-tuning), and the types of encoders employed. In the medical domain, the importance of benchmarking is even more pronounced due to the critical nature of healthcare applications. To the best of our knowledge, no study has yet explored various contrastive learning frameworks in this domain.

\subsection{Broader impact}
\label{sec:limitations}
Our study focuses on benchmarking the performance of multimodal image-text representation learning in the medical domain. We specifically focus on contrastive learning, which is one of the most prominent approaches for learning multimodal data. Our in-depth analysis uncovers key insights into the properties of such methods across various tasks and datasets, laying the groundwork for future advancements in the field.

\subsection{Datasets}
\label{sec:appendix-datasets}

Table \ref{tab:train-data} shows the datasets we used for training all models and each dataset's size and image sub-modality.
The training data is balanced between radiology and histopathology sub-modalities.

\begin{table}[h]
  \caption{Training Datasets}
  \label{tab:train-data}
  \centering
  \begin{tabular}{l|l|l}
  \toprule
    Name     & Image Sub-modality     & Training Split Size \\
    \midrule
    PMC-OA & Radiology and Histopathology  & $1.3$M     \\
    Quilt-1M     & Histopathology & $1.0$M      \\
    MIMIC-IV-CXR     & Radiology       & $369.0$K  \\
    ROCO     & Radiology       & $65.6$K  \\
    \midrule
    SUM     &  & $2.7$M  \\
    \bottomrule
  \end{tabular}
\end{table}

\textbf{MIMIC-CXR}: MIMIC-CXR contains 377,110 images from 65,379 patients, with de-identified free-text
reports describing the images. This dataset is the
largest public chest X-ray dataset, acquired in the emergency department of Beth Israel Deaconess Medical Center in the US. For each patient, there are multiple views and a corresponding report labelled for 13 common radiological conditions using the CheXpert labeller (Irvin et al., 2019) or with “no finding” if no condition is present. Available labels include atelectasis, cardiomegaly, consolidation, edema, enlarged cardiomediastinum, fracture, lung lesion, lung opacity, pleural effusion, pleural other, pneumonia, pneumothorax, support devices, and no finding.

\textbf{Quilt-1M}: Quilt-1M contains more than one million histopathology image-text pairs. This dataset comprises four subsets. The main subset, Quilt, contains $802,144$ image-text pairs sourced from 1,087 hours of education histopathology videos on YouTube. Images and textual captions were automatically extracted from the videos using a mixture of models, including large language models,  handcrafted algorithms, human knowledge databases, and
automatic speech recognition. Three other subsets are sourced from PubMed Open Access Articles, The Large-scale Artificial Intelligence Open Network (LAION-5B), and Twitter data from OpenPath. Image-text pairs which are related to histopathology were extracted from these three datasets and combined with Quilt to constitute more than one million image-text pairs.

\textbf{ROCO}: Radiology Objects in COntext (ROCO) is a set of over 81K radiology image-text pairs with several medical imaging modalities including Computer Tomography, Ultrasound, X-Ray, Fluoroscopy, Positron Emission Tomography,
Mammography, Magnetic Resonance Imaging, Angiography.
ROCO is sourced from open-access biomedical articles on PubMedCentral.

\textbf{PMC-OA}: PMC-OA contains 1.65M image-text pairs encompassing histopathology, radiology, digital camera output, and other modalities. PMC-OA comprises many diagnostic procedures including X-ray, MRI, CT, Fluorescence, Ultrasound, fMRI, ENG, Radioisotope, Endoscope, Mitotic, DOT and PET. The pairs are automatically collected from PubMedCentral's open-access articles via a pipeline proposed in the original paper \citep{lin2023pmc}.

\subsection{Hyperparameter and vision encoder configuration}
\label{sec:appendix-hparam-config}
\begin{figure}
    \centering
    \begin{subfigure}[b]{1\textwidth}
        \centering
        \includegraphics[width=0.7\textwidth]{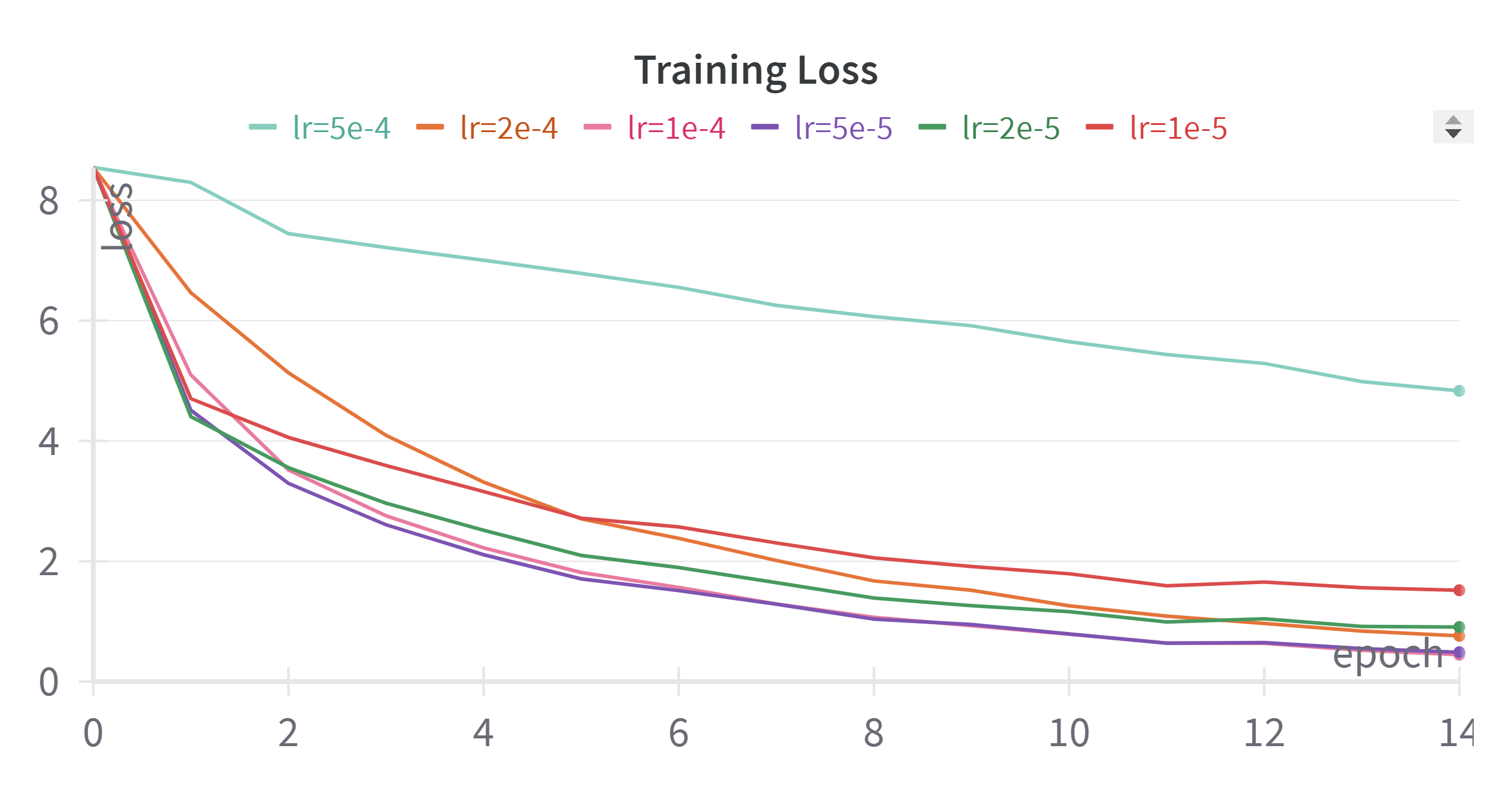}
        \caption{Train loss.}
    \end{subfigure}\\
    \begin{subfigure}[b]{1\textwidth}
        \centering
        \includegraphics[width=0.7\textwidth]{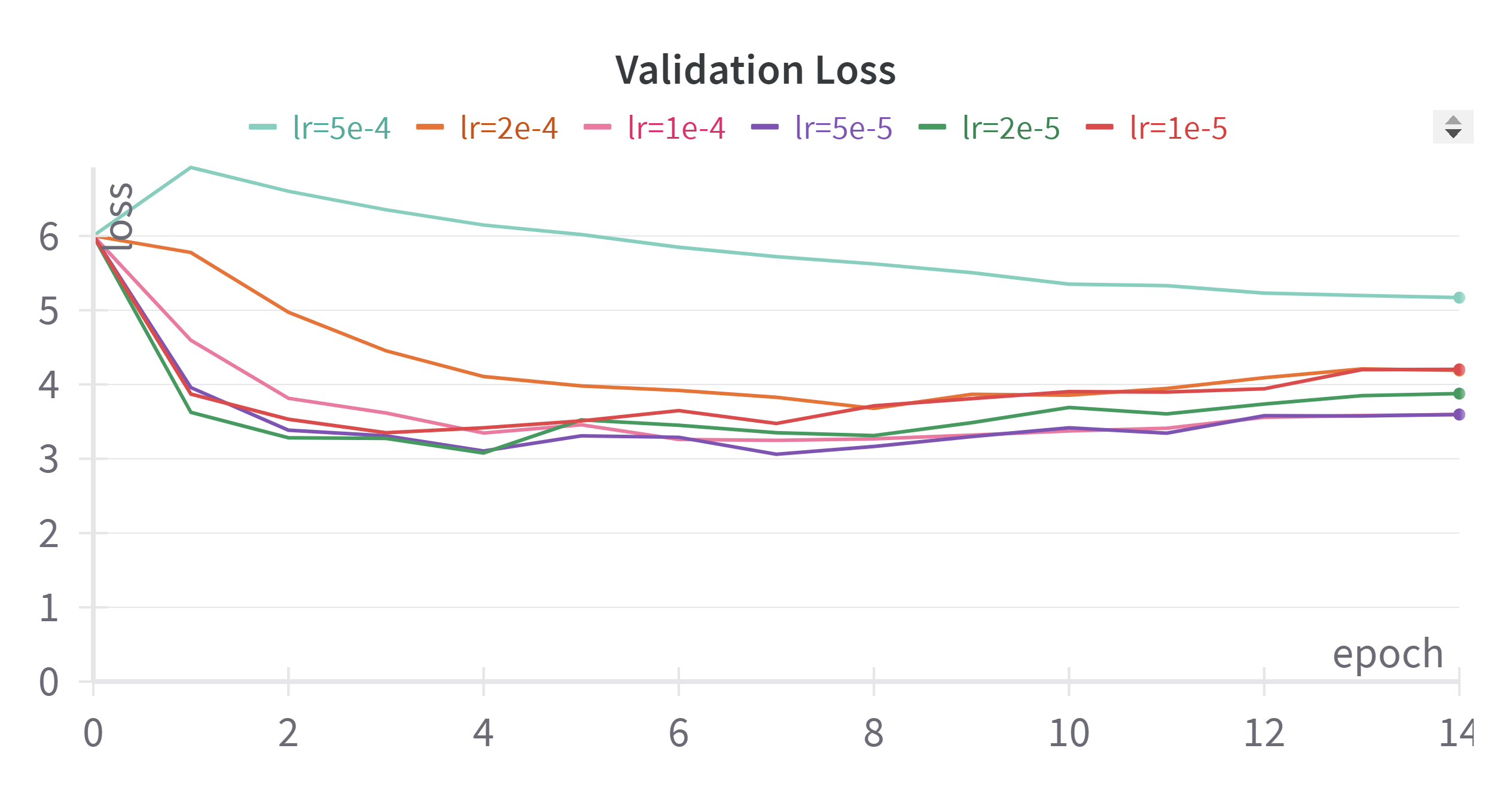}
        \caption{Validation loss.}
    \end{subfigure}
    \caption{Learning rate tuning experiments. A selected model (ViT-B/32 and GPT/77) is trained with various values of the learning rate while training and validation losses are observed. $lr=5e^{-5}$ (purple) yields the least validation loss.}
    \label{fig:lr-tuning}
\end{figure}

All hyperparameters were established using vanilla contrastive learning, achieving the lowest validation loss across four retrieval tasks. Figure \ref{fig:lr-tuning} illustrates an example by plotting the training and validation losses for tuning one of the hyperparameters (learning rate).
We tuned learning rate and batch size by training a selected model (ViT-B/32 and GPT/77) with various values of each hyperparameter and choosing the pair of hyperparameters which yielded the least validation loss.
The chosen hyperparameters are $lr=5e^{-5}$ and batch size of $32$.
Validation was done on the validation split of the PMC-OA dataset.
After selecting the learning rate and batch size, we fixed the text encoder, GPT/77, and tested three different vision encoders to identify the optimal combination for all experiments. The least validation loss and retrieval scores for each encoder are shown in Table \ref{tab:encoder_combination}.
As observed in Table \ref{tab:encoder_combination}, the combination of the vision encoder ViT-B/16 and the GPT/77 text encoder outperformed the other two vision encoders. This combination and the above hyperparameter will be used for all subsequent experiments.

\subsection{Pretraining}
For pretraining, we use a cosine decay learning rate scheduler with no warmup steps. We use the AdamW optimizer \citep{loshchilov2017decoupled} with a weight decay of $0.1$, $\beta_1=0.9$ and $\beta_2=0.999$, along with gradient accumulation with a frequency of 4. Gradient accumulation performs gradient updates after processing 4 batches.
We use a cosine decay learning rate scheduler with no warmup steps.
Our experiments are distributed across four A40 or A100 GPUs.

\subsection{Downstream evaluation}
Tables~
\ref{tab:lp_auc_rq1},
\ref{tab:lp_f1_rq1},
\ref{tab:zsl_auc_rq1},
\ref{tab:zsl_f1_rq1},
\ref{tab:lp_auc_rq2},
\ref{tab:lp_f1_rq2},
\ref{tab:zsl_auc_rq2},
\ref{tab:zsl_f1_rq2},
\ref{tab:lp_auc_rq3},
\ref{tab:lp_f1_rq3},
\ref{tab:zsl_auc_rq3},
\ref{tab:zsl_f1_rq3} show AUC and F1 score of linear probing and zero-shot classification for the three RQs.
Table~\ref{tab:vqa_app_encoders} shows VQA results for encoder selection. Tables~\ref{tab:vqa_app_rq1},~\ref{tab:vqa_app_rq2},~\ref{tab:vqa_app_rq3} show VQA results for the three RQs.

\begin{table}
    \caption{Linear Probing AUC across various medical datasets for models in RQ1.}
    \label{tab:lp_auc_rq1}
    \centering
    \small
    \resizebox{\textwidth}{!}{
    \begin{tabular}{l|c|c|c|c|c||l|c|c|c|c|c}
    \toprule
    \rotatebox{0}{\textbf{Dataset}} & \rotatebox{90}{\textbf{Baseline}} & \rotatebox{90}{\textbf{Image Full Freeze}} & \rotatebox{90}{\textbf{Text Full Freeze}} & \rotatebox{90}{\textbf{Text Partial Freeze}} & \rotatebox{90}{\textbf{Image Partial Freeze}} & \rotatebox{0}{\textbf{Dataset}} & \rotatebox{90}{\textbf{Baseline}} & \rotatebox{90}{\textbf{Image Full Freeze}} & \rotatebox{90}{\textbf{Text Full Freeze}} & \rotatebox{90}{\textbf{Text Partial Freeze}} & \rotatebox{90}{\textbf{Image Partial Freeze}} \\
    \midrule
    {vindr-mammo} & 0.80 & 0.81 & 0.79 & 0.80 & 0.80 & {pathmnist+} & 0.99 & 0.99 & 0.99 & 0.99 & 0.99 \\
    
    {vindr-mammo (mc)} & 0.81 & 0.81 & 0.84 & 0.81 & 0.82 & {dermamnist+} & 0.96 & 0.96 & 0.96 & 0.96 & 0.96 \\
    
    {pad\_ufes\_20} & 0.88 & 0.90 & 0.88 & 0.88 & 0.89 & {octmnist+} & 0.97 & 0.96 & 0.96 & 0.96 & 0.96 \\
    
    {skin\_cancer} & 0.95 & 0.96 & 0.96 & 0.96 & 0.96 & {pneumoniamnist+} & 0.99 & 0.97 & 0.98 & 0.99 & 0.99 \\
    
    {pcam} & 0.95 & 0.91 & 0.94 & 0.95 & 0.95 & {retinamnist+} & 0.86 & 0.83 & 0.85 & 0.86 & 0.87 \\
    
    {nct\_crc\_he\_100k} & 0.98 & 0.99 & 0.99 & 0.99 & 0.99 & {breastmnist+} & 0.88 & 0.88 & 0.89 & 0.89 & 0.88 \\
    
    {lc25000\_lung} & 0.99 & 1.00 & 1.00 & 1.00 & 1.00 & {bloodmnist+} & 0.99 & 0.79 & 0.91 & 0.91 & 0.91 \\
    
    {lc25000\_colon} & 1.00 & 1.00 & 1.00 & 1.00 & 1.00 & {tissuemnist+} & 0.91 & 0.89 & 0.91 & 0.91 & 0.91 \\
    
    {bach} & 0.98 & 0.90 & 0.99 & 0.97 & 0.98 & {organamnist+} & 0.99 & 0.99 & 0.99 & 1.00 & 0.99 \\
    
    {sicap} & 0.91 & 0.88 & 0.88 & 0.91 & 0.98 & {organcmnist+} & 0.99 & 0.99 & 0.99 & 0.99 & 0.99 \\
    
    {chestmnist+} & 0.78 & 0.75 & 0.77 & 0.79 & 0.80 & {organsmnist+} & 0.97 & 0.97 & 0.97 & 0.98 & 0.97 \\
    \bottomrule
    \end{tabular}
    }
\end{table}

\begin{table}
    \caption{Linear Probing F1-score across various medical datasets for models in RQ1.}
    \label{tab:lp_f1_rq1}
    \centering
    \small
    \resizebox{\textwidth}{!}{
    \begin{tabular}{l|c|c|c|c|c||l|c|c|c|c|c}
    \toprule
    \rotatebox{0}{\textbf{Dataset}} & \rotatebox{90}{\textbf{Baseline}} & \rotatebox{90}{\textbf{Image Full Freeze}} & \rotatebox{90}{\textbf{Text Full Freeze}} & \rotatebox{90}{\textbf{Text Partial Freeze}} & \rotatebox{90}{\textbf{Image Partial Freeze}} & \rotatebox{0}{\textbf{Dataset}} & \rotatebox{90}{\textbf{Baseline}} & \rotatebox{90}{\textbf{Image Full Freeze}} & \rotatebox{90}{\textbf{Text Full Freeze}} & \rotatebox{90}{\textbf{Text Partial Freeze}} & \rotatebox{90}{\textbf{Image Partial Freeze}} \\
    \midrule
    {vindr-mammo} & 0.03 & 0.04 & 0.15 & 0.07 & 0.04 & {pathmnist+} & 0.91 & 0.88 & 0.99 & 0.92 & 0.99 \\
    
    {vindr-mammo (mc)} & 0.18 & 0.20 & 0.18 & 0.19 & 0.22 & {dermamnist+} & 0.68 & 0.68 & 0.70 & 0.70 & 0.71 \\
    
    {pad\_ufes\_20} & 0.54 & 0.58 & 0.52 & 0.58 & 0.59 & {octmnist+} & 0.73 & 0.68 & 0.70 & 0.71 & 0.71 \\
    
    {skin\_cancer} & 0.68 & 0.66 & 0.67 & 0.67 & 0.66 & {pneumoniamnist+} & 0.91 & 0.85 & 0.88 & 0.90 & 0.93 \\
    
    {pcam} & 0.88 & 0.82 & 0.85 & 0.88 & 0.87 & {retinamnist+} & 0.49 & 0.49 & 0.48 & 0.49 & 0.50 \\
    
    {nct\_crc\_he\_100k} & 0.90 & 0.86 & 0.88 & 0.91 & 0.90 & {breastmnist+} & 0.84 & 0.79 & 0.78 & 0.77 & 0.78 \\
    
    {lc25000\_lung} & 0.99 & 0.98 & 1.00 & 1.00 & 1.00 & {bloodmnist+} & 0.97 & 0.89 & 0.91 & 0.91 & 0.91 \\
    
    {lc25000\_colon} & 1.00 & 1.00 & 1.00 & 1.00 & 1.00 & {tissuemnist+} & 0.54 & 0.49 & 0.54 & 0.54 & 0.52 \\
    
    {bach} & 0.88 & 0.67 & 0.99 & 0.97 & 0.98 & {organamnist+} & 0.92 & 0.89 & 0.99 & 1.00 & 0.99 \\
    
    {sicap} & 0.67 & 0.61 & 0.88 & 0.91 & 0.98 & {organcmnist+} & 0.85 & 0.81 & 0.99 & 0.99 & 0.99 \\
    
    {chestmnist+} & 0.08 & 0.02 & 0.06 & 0.07 & 0.09 & {organsmnist+} & 0.73 & 0.70 & 0.99 & 0.98 & 0.97 \\
    \bottomrule
    \end{tabular}
    }
\end{table}

\begin{table}
    \caption{Zero-shot classification AUC across various medical datasets for models in RQ1.}
    \label{tab:zsl_auc_rq1}
    \centering
    \small
    \resizebox{\textwidth}{!}{
    \begin{tabular}{l|c|c|c|c|c||l|c|c|c|c|c}
    \toprule
    \rotatebox{0}{\textbf{Dataset}} & \rotatebox{90}{\textbf{Baseline}} & \rotatebox{90}{\textbf{Image Full Freeze}} & \rotatebox{90}{\textbf{Text Full Freeze}} & \rotatebox{90}{\textbf{Text Partial Freeze}} & \rotatebox{90}{\textbf{Image Partial Freeze}} & \rotatebox{0}{\textbf{Dataset}} & \rotatebox{90}{\textbf{Baseline}} & \rotatebox{90}{\textbf{Image Full Freeze}} & \rotatebox{90}{\textbf{Text Full Freeze}} & \rotatebox{90}{\textbf{Text Partial Freeze}} & \rotatebox{90}{\textbf{Image Partial Freeze}} \\
    \midrule
    {vindr-mammo} & 0.50 & 0.43 & 0.48 & 0.50 & 0.50 & {pathmnist+} & 0.88 & 0.79 & 0.81 & 0.83 & 0.86 \\
    {vindr-mammo (mc)} & 0.44 & 0.44 & 0.47 & 0.43 & 0.38 & {dermamnist+} & 0.60 & 0.50 & 0.64 & 0.63 & 0.64 \\
    {pad\_ufes\_20} & 0.59 & 0.54 & 0.61 & 0.55 & 0.66 & {octmnist+} & 0.68 & 0.55 & 0.73 & 0.72 & 0.65 \\
    {skin\_cancer} & 0.57 & 0.48 & 0.66 & 0.72 & 0.63 & {pneumoniamnist+} & 0.94 & 0.52 & 0.93 & 0.84 & 0.85 \\
    {pcam} & 0.71 & 0.62 & 0.84 & 0.57 & 0.78 & {retinamnist+} & 0.59 & 0.48 & 0.41 & 0.47 & 0.56 \\
    {nct\_crc\_he\_100k} & 0.95 & 0.78 & 0.86 & 0.91 & 0.92 & {breastmnist+} & 0.53 & 0.56 & 0.44 & 0.62 & 0.46 \\
    {lc25000\_lung} & 0.97 & 0.59 & 0.98 & 0.97 & 0.94 & {bloodmnist+} & 0.65 & 0.56 & 0.67 & 0.56 & 0.59 \\
    {lc25000\_colon} & 0.99 & 0.91 & 0.88 & 0.99 & 1.00 & {tissuemnist+} & 0.45 & 0.41 & 0.50 & 0.50 & 0.48 \\
    {bach} & 0.82 & 0.62 & 0.66 & 0.73 & 0.79 & {organamnist+} & 0.85 & 0.73 & 0.82 & 0.84 & 0.85 \\
    {sicap} & 0.71 & 0.50 & 0.72 & 0.75 & 0.79 & {organcmnist+} & 0.78 & 0.66 & 0.77 & 0.77 & 0.78 \\
    {chestmnist+} & 0.50 & 0.52 & 0.47 & 0.50 & 0.50 & {organsmnist+} & 0.81 & 0.63 & 0.78 & 0.78 & 0.80 \\
    \bottomrule
    \end{tabular}
    }
\end{table}

\begin{table}
    \caption{Zero-shot classification F1-score across various medical datasets for models in RQ1.}
    \label{tab:zsl_f1_rq1}
    \centering
    \small
    \resizebox{\textwidth}{!}{
    \begin{tabular}{l|c|c|c|c|c||l|c|c|c|c|c}
    \toprule
    \rotatebox{0}{\textbf{Dataset}} & \rotatebox{90}{\textbf{Baseline}} & \rotatebox{90}{\textbf{Image Full Freeze}} & \rotatebox{90}{\textbf{Text Full Freeze}} & \rotatebox{90}{\textbf{Text Partial Freeze}} & \rotatebox{90}{\textbf{Image Partial Freeze}} & \rotatebox{0}{\textbf{Dataset}} & \rotatebox{90}{\textbf{Baseline}} & \rotatebox{90}{\textbf{Image Full Freeze}} & \rotatebox{90}{\textbf{Text Full Freeze}} & \rotatebox{90}{\textbf{Text Partial Freeze}} & \rotatebox{90}{\textbf{Image Partial Freeze}} \\
    \midrule
    {vindr-mammo} & 0.02 & 0.02 & 0.02 & 0.02 & 0.02 & {pathmnist+} & 0.44 & 0.37 & 0.27 & 0.38 & 0.43 \\
    {vindr-mammo (mc)} & 0.01 & 0.00 & 0.09 & 0.00 & 0.04 & {dermamnist+} & 0.07 & 0.05 & 0.09 & 0.07 & 0.06 \\
    {pad\_ufes\_20} & 0.12 & 0.08 & 0.21 & 0.07 & 0.15 & {octmnist+} & 0.23 & 0.10 & 0.26 & 0.25 & 0.24 \\
    {skin\_cancer} & 0.07 & 0.04 & 0.11 & 0.08 & 0.03 & {pneumoniamnist+} & 0.52 & 0.38 & 0.39 & 0.71 & 0.77 \\
    {pcam} & 0.66 & 0.57 & 0.69 & 0.54 & 0.65 & {retinamnist+} & 0.05 & 0.14 & 0.09 & 0.08 & 0.12 \\
    {nct\_crc\_he\_100k} & 0.54 & 0.33 & 0.41 & 0.56 & 0.53 & {breastmnist+} & 0.52 & 0.55 & 0.22 & 0.47 & 0.47 \\
    {lc25000\_lung} & 0.73 & 0.26 & 0.90 & 0.85 & 0.77 & {bloodmnist+} & 0.03 & 0.56 & 0.10 & 0.47 & 0.59 \\
    {lc25000\_colon} & 0.95 & 0.82 & 0.41 & 0.93 & 0.94 & {tissuemnist+} & 0.03 & 0.04 & 0.09 & 0.04 & 0.03 \\
    {bach} & 0.54 & 0.23 & 0.32 & 0.34 & 0.42 & {organamnist+} & 0.28 & 0.13 & 0.20 & 0.21 & 0.23 \\
    {sicap} & 0.40 & 0.12 & 0.30 & 0.35 & 0.32 & {organcmnist+} & 0.19 & 0.11 & 0.20 & 0.19 & 0.16 \\
    {chestmnist+} & 0.09 & 0.10 & 0.10 & 0.10 & 0.10 & {organsmnist+} & 0.21 & 0.09 & 0.19 & 0.19 & 0.18 \\
    \bottomrule
    \end{tabular}
    }
\end{table}

\begin{table}
    \caption{Linear probing AUC across various medical datasets for models in RQ2.}
    \label{tab:lp_auc_rq2}
    \centering
    \small
    \begin{tabular}{l|c|c||l|c|c}
    \toprule
    \textbf{Dataset} & \rotatebox{90}{\textbf{Augmented CL}} & \rotatebox{90}{\textbf{Masked CL}} & \textbf{Dataset} & \rotatebox{90}{\textbf{Augmented CL}} & \rotatebox{90}{\textbf{Masked CL}} \\
    \midrule
    vindr-mammo & 0.75 & 0.76 & pathmnist+ & 0.99 & 0.99 \\
    vindr-mammo (multiclass) & 0.78 & 0.81 & dermamnist+ & 0.90 & 0.90 \\
    pad\_ufes\_20 & 0.80 & 0.82 & octmnist+ & 0.95 & 0.93 \\
    skin\_cancer & 0.90 & 0.90 & pneumoniamnist+ & 0.97 & 0.96 \\
    pcam & 0.93 & 0.94 & retinamnist+ & 0.84 & 0.78 \\
    nct\_crc\_he\_100k & 0.99 & 0.99 & breastmnist+ & 0.76 & 0.82 \\
    lc25000\_lung & 0.97 & 0.97 & bloodmnist+ & 0.98 & 0.98 \\
    lc25000\_colon & 1.00 & 1.00 & tissuemnist+ & 0.89 & 0.89 \\
    bach & 0.97 & 0.98 & organamnist+ & 0.99 & 0.99 \\
    sicap & 0.91 & 0.91 & organcmnist+ & 0.99 & 0.99 \\
    chestmnist+ & 0.79 & 0.79 & organsmnist+ & 0.98 & 0.98 \\
    \bottomrule
    \end{tabular}
\end{table}

\begin{table}
    \caption{Linear probing F1-score across various medical datasets for models in RQ2.}
    \label{tab:lp_f1_rq2}
    \centering
    \small
    \begin{tabular}{l|c|c||l|c|c}
    \toprule
    \textbf{Dataset} & \rotatebox{90}{\textbf{Augmented CL}} & \rotatebox{90}{\textbf{Masked CL}} & \textbf{Dataset} & \rotatebox{90}{\textbf{Augmented CL}} & \rotatebox{90}{\textbf{Masked CL}} \\
    \midrule
    vindr-mammo & 0.00 & 0.00 & pathmnist+ & 0.91 & 0.92 \\
    vindr-mammo (multiclass) & 0.10 & 0.10 & dermamnist+ & 0.39 & 0.42 \\
    pad\_ufes\_20 & 0.41 & 0.36 & octmnist+ & 0.64 & 0.51 \\
    skin\_cancer & 0.42 & 0.40 & pneumoniamnist+ & 0.84 & 0.84 \\
    pcam & 0.85 & 0.86 & retinamnist+ & 0.33 & 0.30 \\
    nct\_crc\_he\_100k & 0.50 & 0.42 & breastmnist+ & 0.42 & 0.42 \\
    lc25000\_lung & 0.97 & 0.97 & bloodmnist+ & 0.84 & 0.84 \\
    lc25000\_colon & 1.00 & 1.00 & tissuemnist+ & 0.45 & 0.45 \\
    bach & 0.55 & 0.54 & organamnist+ & 0.85 & 0.84 \\
    sicap & 0.67 & 0.60 & organcmnist+ & 0.65 & 0.64 \\
    chestmnist+ & 0.01 & 0.01 & organsmnist+ & 0.64 & 0.64 \\
    \bottomrule
    \end{tabular}
\end{table}

\begin{table}
    \caption{Zero-shot classification AUC across various medical datasets for models in RQ2.}
    \label{tab:zsl_auc_rq2}
    \centering
    \small
    \begin{tabular}{l|c|c||l|c|c}
    \toprule
    \textbf{Dataset} & \rotatebox{90}{\textbf{Augmented CL}} & \rotatebox{90}{\textbf{Masked CL}} & \textbf{Dataset} & \rotatebox{90}{\textbf{Augmented CL}} & \rotatebox{90}{\textbf{Masked CL}} \\
    \midrule
    vindr-mammo & 0.42 & 0.51 & pathmnist+ & 0.78 & 0.84 \\
    vindr-mammo (multiclass) & 0.54 & 0.56 & dermamnist+ & 0.44 & 0.49 \\
    pad\_ufes\_20 & 0.51 & 0.48 & octmnist+ & 0.50 & 0.62 \\
    skin\_cancer & 0.47 & 0.53 & pneumoniamnist+ & 0.20 & 0.65 \\
    pcam & 0.82 & 0.76 & retinamnist+ & 0.47 & 0.48 \\
    nct\_crc\_he\_100k & 0.89 & 0.87 & breastmnist+ & 0.57 & 0.58 \\
    lc25000\_lung & 0.98 & 0.96 & bloodmnist+ & 0.48 & 0.44 \\
    lc25000\_colon & 0.92 & 0.96 & tissuemnist+ & 0.49 & 0.50 \\
    bach & 0.60 & 0.65 & organamnist+ & 0.79 & 0.67 \\
    sicap & 0.59 & 0.64 & organcmnist+ & 0.72 & 0.69 \\
    chestmnist+ & 0.41 & 0.44 & organsmnist+ & 0.75 & 0.72 \\
    \bottomrule
    \end{tabular}
\end{table}

\begin{table}
    \caption{Zero-shot classification F1-score across various medical datasets for models in RQ2.}
    \label{tab:zsl_f1_rq2}
    \centering
    \small
    \begin{tabular}{l|c|c||l|c|c}
    \toprule
    \textbf{Dataset} & \rotatebox{90}{\textbf{Augmented CL}} & \rotatebox{90}{\textbf{Masked CL}} & \textbf{Dataset} & \rotatebox{90}{\textbf{Augmented CL}} & \rotatebox{90}{\textbf{Masked CL}} \\
    \midrule
    vindr-mammo & 0.02 & 0.02 & pathmnist+ & 0.30 & 0.32 \\
    vindr-mammo (multiclass) & 0.07 & 0.04 & dermamnist+ & 0.12 & 0.03 \\
    pad\_ufes\_20 & 0.09 & 0.07 & octmnist+ & 0.10 & 0.18 \\
    skin\_cancer & 0.13 & 0.12 & pneumoniamnist+ & 0.20 & 0.18 \\
    pcam & 0.74 & 0.69 & retinamnist+ & 0.05 & 0.05 \\
    nct\_crc\_he\_100k & 0.44 & 0.37 & breastmnist+ & 0.25 & 0.50 \\
    lc25000\_lung & 0.89 & 0.86 & bloodmnist+ & 0.09 & 0.12 \\
    lc25000\_colon & 0.67 & 0.88 & tissuemnist+ & 0.05 & 0.12 \\
    bach & 0.39 & 0.29 & organamnist+ & 0.13 & 0.12 \\
    sicap & 0.23 & 0.23 & organcmnist+ & 0.12 & 0.12 \\
    chestmnist+ & 0.10 & 0.10 & organsmnist+ & 0.12 & 0.13 \\
    \bottomrule
    \end{tabular}
\end{table}

\begin{table}
    \caption{Linear probing AUC across various medical datasets for models in RQ3.}
    \label{tab:lp_auc_rq3}
    \centering
    \small
    \begin{tabular}{l|c|c||l|c|c}
    \toprule
    \textbf{Dataset} & \rotatebox{90}{\textbf{Image Captioning}} & \rotatebox{90}{\textbf{Fast CL}} & \textbf{Dataset} & \rotatebox{90}{\textbf{Image Captioning}} & \rotatebox{90}{\textbf{Fast CL}} \\
    \midrule
    vindr-mammo & 0.81 & 0.83 & pathmnist+ & 0.99 & 0.99 \\
    vindr-mammo (multiclass) & 0.83 & 0.83 & dermamnist+ & 0.95 & 0.96 \\
    pad\_ufes\_20 & 0.90 & 0.87 & octmnist+ & 0.96 & 0.96 \\
    skin\_cancer & 0.95 & 0.96 & pneumoniamnist+ & 0.98 & 0.98 \\
    pcam & 0.96 & 0.91 & retinamnist+ & 0.85 & 0.86 \\
    nct\_crc\_he\_100k & 0.99 & 0.99 & breastmnist+ & 0.87 & 0.90 \\
    lc25000\_lung & 1.00 & 1.00 & bloodmnist+ & 1.00 & 1.00 \\
    lc25000\_colon & 1.00 & 1.00 & tissuemnist+ & 0.91 & 0.91 \\
    bach & 0.97 & 0.96 & organamnist+ & 0.99 & 0.99 \\
    sicap & 0.92 & 0.92 & organcmnist+ & 0.99 & 0.99 \\
    chestmnist+ & 0.96 & 0.96 & organsmnist+ & 0.98 & 0.98 \\
    \bottomrule
    \end{tabular}
\end{table}

\begin{table}
    \caption{Linear probing F1-score across various medical datasets for models in RQ3.}
    \label{tab:lp_f1_rq3}
    \centering
    \small
    \begin{tabular}{l|c|c||l|c|c}
    \toprule
    \textbf{Dataset} & \rotatebox{90}{\textbf{Image Captioning}} & \rotatebox{90}{\textbf{Fast CL}} & \textbf{Dataset} & \rotatebox{90}{\textbf{Image Captioning}} & \rotatebox{90}{\textbf{Fast CL}} \\
    \midrule
    vindr-mammo & 0.04 & 0.04 & pathmnist+ & 0.92 & 0.92 \\
    vindr-mammo (multiclass) & 0.19 & 0.19 & dermamnist+ & 0.39 & 0.39 \\
    pad\_ufes\_20 & 0.57 & 0.53 & octmnist+ & 0.42 & 0.39 \\
    skin\_cancer & 0.65 & 0.66 & pneumoniamnist+ & 0.84 & 0.84 \\
    pcam & 0.88 & 0.88 & retinamnist+ & 0.50 & 0.50 \\
    nct\_crc\_he\_100k & 0.61 & 0.59 & breastmnist+ & 0.79 & 0.77 \\
    lc25000\_lung & 1.00 & 0.53 & bloodmnist+ & 0.49 & 0.53 \\
    lc25000\_colon & 0.95 & 0.53 & tissuemnist+ & 0.54 & 0.54 \\
    bach & 0.83 & 0.80 & organamnist+ & 0.90 & 0.88 \\
    sicap & 0.65 & 0.62 & organcmnist+ & 0.73 & 0.72 \\
    chestmnist+ & 0.06 & 0.07 & organsmnist+ & 0.77 & 0.77 \\
    \bottomrule
    \end{tabular}
\end{table}

\begin{table}
    \caption{Zero-shot classification AUC across various medical datasets for models in RQ3.}
    \label{tab:zsl_auc_rq3}
    \centering
    \small
    \begin{tabular}{l|c|c||l|c|c}
    \toprule
    \textbf{Dataset} & \rotatebox{90}{\textbf{Image Captioning}} & \rotatebox{90}{\textbf{Fast CL}} & \textbf{Dataset} & \rotatebox{90}{\textbf{Image Captioning}} & \rotatebox{90}{\textbf{Fast CL}} \\
    \midrule
    vindr-mammo & 0.50 & 0.50 & pathmnist+ & 0.84 & 0.85 \\
    vindr-mammo (multiclass) & 0.45 & 0.43 & dermamnist+ & 0.60 & 0.60 \\
    pad\_ufes\_20 & 0.66 & 0.57 & octmnist+ & 0.57 & 0.56 \\
    skin\_cancer & 0.67 & 0.59 & pneumoniamnist+ & 0.90 & 0.80 \\
    pcam & 0.61 & 0.61 & retinamnist+ & 0.50 & 0.50 \\
    nct\_crc\_he\_100k & 0.91 & 0.91 & breastmnist+ & 0.71 & 0.53 \\
    lc25000\_lung & 0.95 & 0.91 & bloodmnist+ & 0.60 & 0.66 \\
    lc25000\_colon & 0.99 & 0.99 & tissuemnist+ & 0.53 & 0.50 \\
    bach & 0.72 & 0.72 & organamnist+ & 0.78 & 0.81 \\
    sicap & 0.70 & 0.67 & organcmnist+ & 0.76 & 0.74 \\
    chestmnist+ & 0.50 & 0.50 & organsmnist+ & 0.78 & 0.75 \\
    \bottomrule
    \end{tabular}
\end{table}

\begin{table}
    \caption{Zero-shot classification F1-score across various medical datasets for models in RQ3.}
    \label{tab:zsl_f1_rq3}
    \centering
    \small
    \begin{tabular}{l|c|c||l|c|c}
    \toprule
    \textbf{Dataset} & \rotatebox{90}{\textbf{Image Captioning}} & \rotatebox{90}{\textbf{Fast CL}} & \textbf{Dataset} & \rotatebox{90}{\textbf{Image Captioning}} & \rotatebox{90}{\textbf{Fast CL}} \\
    \midrule
    vindr-mammo & 0.02 & 0.02 & pathmnist+ & 0.51 & 0.51 \\
    vindr-mammo (multiclass) & 0.10 & 0.01 & dermamnist+ & 0.05 & 0.04 \\
    pad\_ufes\_20 & 0.18 & 0.12 & octmnist+ & 0.12 & 0.10 \\
    skin\_cancer & 0.16 & 0.13 & pneumoniamnist+ & 0.78 & 0.71 \\
    pcam & 0.47 & 0.45 & retinamnist+ & 0.14 & 0.14 \\
    nct\_crc\_he\_100k & 0.53 & 0.53 & breastmnist+ & 0.42 & 0.21 \\
    lc25000\_lung & 0.62 & 0.53 & bloodmnist+ & 0.49 & 0.53 \\
    lc25000\_colon & 0.95 & 0.53 & tissuemnist+ & 0.08 & 0.12 \\
    bach & 0.32 & 0.29 & organamnist+ & 0.13 & 0.12 \\
    sicap & 0.29 & 0.29 & organcmnist+ & 0.13 & 0.13 \\
    chestmnist+ & 0.10 & 0.10 & organsmnist+ & 0.13 & 0.13 \\
    \bottomrule
    \end{tabular}
\end{table}

\begin{table}[]
\centering
\caption{Visual question answering Accuracy score across two medical datasets for models in Table \ref{tab:encoder_combination}.}
\label{tab:vqa_app_encoders}
\begin{tabular}{c|rrr|rrrr}
\hline
\multirow{2}{*}{\textbf{Vision encoders}} & \multicolumn{3}{c|}{\textbf{VQARAD}}             & \multicolumn{4}{c}{\textbf{PathVQA}}                              \\
 &
  \multicolumn{1}{c}{Open-ended} &
  \multicolumn{1}{c}{Close-ended} &
  \multicolumn{1}{c|}{Overall} &
  \multicolumn{1}{c}{Yes/No} &
  \multicolumn{1}{c}{Number} &
  \multicolumn{1}{c}{Other} &
  \multicolumn{1}{c}{Overall} \\ \hline
RN50                                      & \textbf{45.81} & 73.53          & 62.53          & 82.31          & \textbf{33.33} & \textbf{10.23} & 46.44          \\
ViT-B/16                                  & 42.46          & 74.26          & 61.64          & \textbf{83.19} & 27.78          & 9.99           & \textbf{46.75} \\
ViT-B/32                                  & 44.13          & \textbf{75.37} & \textbf{62.97} & 82.63          & 27.78          & 9.7            & 46.33          \\ \hline
\end{tabular}
\end{table}

\begin{table}[]
\centering
\caption{Visual question answering Accuracy score across two medical datasets for models in RQ1.}
\label{tab:vqa_app_rq1}
\begin{tabular}{c|ccc|cccc}
\hline
\multirow{2}{*}{\textbf{Models}} & \multicolumn{3}{c|}{\textbf{VQARAD}}            & \multicolumn{4}{c}{\textbf{PathVQA}}           \\
                     & Open-ended & Close-ended & Overall & Yes/No         & Number & Other          & Overall        \\ \hline
Baseline             & 42.46      & 74.26       & 61.64   & \textbf{83.19} & 27.78  & 9.99           & \textbf{46.75} \\
Image Full Freeze                & 38.55          & 73.9          & 59.87          & 82.1 & \textbf{33.33} & \textbf{10.11} & 46.28 \\
Text Full Freeze     & 40.78      & 72.47       & 59.87   & 82.96          & 27.78  & \textbf{10.11} & 46.69          \\
Image Partial Freeze & 41.9       & 74.26       & 61.42   & 82.54          & 27.78  & 9.25           & 46.06          \\
Text Partial Freeze              & \textbf{43.58} & \textbf{76.1} & \textbf{63.19} & 82.6 & \textbf{33.33} & 9.67           & 46.31 \\ \hline
\end{tabular}
\end{table}

\begin{table}[]
\centering
\caption{Visual question answering Accuracy score across two medical datasets for models in RQ2.}
\label{tab:vqa_app_rq2}
\begin{tabular}{l|rrr|rrrr}
\hline
\multicolumn{1}{c|}{\multirow{2}{*}{\textbf{Models}}} & \multicolumn{3}{c|}{\textbf{VQARAD}}          & \multicolumn{4}{c}{\textbf{PathVQA}}                              \\
\multicolumn{1}{c|}{} &
  \multicolumn{1}{c}{Open-ended} &
  \multicolumn{1}{c}{Close-ended} &
  \multicolumn{1}{c|}{Overall} &
  \multicolumn{1}{c}{Yes/No} &
  \multicolumn{1}{c}{Number} &
  \multicolumn{1}{c}{Other} &
  \multicolumn{1}{c}{Overall} \\ \hline
Baseline                                              & 42.46          & 74.26       & 61.64          & \textbf{83.19} & 27.78          & 9.99           & 46.75          \\
Augmented CL                                          & 46.37          & \textbf{75} & 63.64          & 82.54          & \textbf{33.33} & 9.61           & 46.25          \\
Masked CL                                             & \textbf{49.16} & 74.63       & \textbf{64.52} & 83.13          & \textbf{33.33} & \textbf{10.23} & \textbf{46.86} \\ \hline
\end{tabular}
\end{table}

\begin{table}[]
\centering
\caption{Visual question answering Accuracy score across two medical datasets for models in RQ3.}
\label{tab:vqa_app_rq3}
\begin{tabular}{l|rrr|rrrr}
\hline
\multicolumn{1}{c|}{\multirow{2}{*}{\textbf{Models}}} &
  \multicolumn{3}{c|}{\textbf{VQARAD}} &
  \multicolumn{4}{c}{\textbf{PathVQA}} \\
\multicolumn{1}{c|}{} &
  \multicolumn{1}{c}{Open-ended} &
  \multicolumn{1}{c}{Close-ended} &
  \multicolumn{1}{c|}{Overall} &
  \multicolumn{1}{c}{Yes/No} &
  \multicolumn{1}{c}{Number} &
  \multicolumn{1}{c}{Other} &
  \multicolumn{1}{c}{Overall} \\ \hline
Baseline &
  \textbf{42.46} &
  74.26 &
  61.64 &
  \textbf{83.19} &
  27.78 &
  9.99 &
  46.75 \\
FastCL 50\% &
  40.22 &
  75.74 &
  61.64 &
  82.31 &
  27.78 &
  9.88 &
  46.25 \\
FastCL 75\% &
  \textbf{42.46} &
  72.79 &
  60.75 &
  82.87 &
  38.89 &
  10.17 &
  46.71 \\
Image captioning &
  41.34 &
  \textbf{76.47} &
  \textbf{62.53} &
  82.6 &
  \textbf{61.11} &
  \textbf{11.84} &
  \textbf{47.46} \\ \hline
\end{tabular}
\end{table}

%% file: main.bbl
\begin{thebibliography}{}

\bibitem[Acosta et~al., 2022]{acosta2022multimodal}
Acosta, J.~N., Falcone, G.~J., Rajpurkar, P., and Topol, E.~J. (2022).
\newblock Multimodal biomedical ai.
\newblock {\em Nature Medicine}, 28(9):1773--1784.

\bibitem[Alsheh~Ali et~al., 2019]{alsheh2019association}
Alsheh~Ali, M., Czene, K., Hall, P., and Humphreys, K. (2019).
\newblock Association of microcalcification clusters with short-term invasive breast cancer risk and breast cancer risk factors.
\newblock {\em Scientific reports}, 9(1):14604.

\bibitem[Baliah et~al., 2023]{baliah2023exploring}
Baliah, S., Maani, F.~A., Sanjeev, S., and Khan, M.~H. (2023).
\newblock Exploring the transfer learning capabilities of clip in domain generalization for diabetic retinopathy.
\newblock In {\em International Workshop on Machine Learning in Medical Imaging}, pages 444--453. Springer.

\bibitem[Boecking et~al., 2022]{boecking2022making}
Boecking, B., Usuyama, N., Bannur, S., Castro, D.~C., Schwaighofer, A., Hyland, S., Wetscherek, M., Naumann, T., Nori, A., Alvarez-Valle, J., et~al. (2022).
\newblock Making the most of text semantics to improve biomedical vision--language processing.
\newblock In {\em European conference on computer vision}, pages 1--21. Springer.

\bibitem[Chaitanya et~al., 2020]{chaitanya2020contrastive}
Chaitanya, K., Erdil, E., Karani, N., and Konukoglu, E. (2020).
\newblock Contrastive learning of global and local features for medical image segmentation with limited annotations.
\newblock {\em Advances in neural information processing systems}, 33:12546--12558.

\bibitem[Chen et~al., 2024]{chen2024benchmarking}
Chen, S., Gu, J., Han, Z., Ma, Y., Torr, P., and Tresp, V. (2024).
\newblock Benchmarking robustness of adaptation methods on pre-trained vision-language models.
\newblock {\em Advances in Neural Information Processing Systems}, 36.

\bibitem[Chen et~al., 2020]{chen2020simple}
Chen, T., Kornblith, S., Norouzi, M., and Hinton, G. (2020).
\newblock A simple framework for contrastive learning of visual representations.
\newblock In {\em International conference on machine learning}, pages 1597--1607. PMLR.

\bibitem[Cui et~al., 2022]{cui2022democratizing}
Cui, Y., Zhao, L., Liang, F., Li, Y., and Shao, J. (2022).
\newblock Democratizing contrastive language-image pre-training: A clip benchmark of data, model, and supervision.
\newblock {\em arXiv preprint arXiv:2203.05796}.

\bibitem[Dattakumar and Jagadeesh, 2003]{dattakumar2003review}
Dattakumar, R. and Jagadeesh, R. (2003).
\newblock A review of literature on benchmarking.
\newblock {\em Benchmarking: An International Journal}, 10(3):176--209.

\bibitem[Do et~al., 2021]{do2021multiple}
Do, T., Nguyen, B.~X., Tjiputra, E., Tran, M., Tran, Q.~D., and Nguyen, A. (2021).
\newblock Multiple meta-model quantifying for medical visual question answering.
\newblock In {\em Medical Image Computing and Computer Assisted Intervention--MICCAI 2021: 24th International Conference, Strasbourg, France, September 27--October 1, 2021, Proceedings, Part V 24}, pages 64--74. Springer.

\bibitem[Dosovitskiy et~al., 2020]{dosovitskiy2020image}
Dosovitskiy, A., Beyer, L., Kolesnikov, A., Weissenborn, D., Zhai, X., Unterthiner, T., Dehghani, M., Minderer, M., Heigold, G., Gelly, S., et~al. (2020).
\newblock An image is worth 16x16 words: Transformers for image recognition at scale.
\newblock In {\em International Conference on Learning Representations}.

\bibitem[Driess et~al., 2023]{driess2023palm}
Driess, D., Xia, F., Sajjadi, M.~S., Lynch, C., Chowdhery, A., Ichter, B., Wahid, A., Tompson, J., Vuong, Q., Yu, T., et~al. (2023).
\newblock Palm-e: An embodied multimodal language model.
\newblock In {\em International Conference on Machine Learning}, pages 8469--8488. PMLR.

\bibitem[Eslami et~al., 2021]{eslami2021does}
Eslami, S., de~Melo, G., and Meinel, C. (2021).
\newblock Does clip benefit visual question answering in the medical domain as much as it does in the general domain?
\newblock {\em arXiv preprint arXiv:2112.13906}.

\bibitem[Girdhar et~al., 2023]{girdhar2023imagebind}
Girdhar, R., El-Nouby, A., Liu, Z., Singh, M., Alwala, K.~V., Joulin, A., and Misra, I. (2023).
\newblock Imagebind: One embedding space to bind them all.
\newblock In {\em Proceedings of the IEEE/CVF Conference on Computer Vision and Pattern Recognition}, pages 15180--15190.

\bibitem[He et~al., 2016]{he2016deep}
He, K., Zhang, X., Ren, S., and Sun, J. (2016).
\newblock Deep residual learning for image recognition.
\newblock In {\em Proceedings of the IEEE conference on computer vision and pattern recognition}, pages 770--778.

\bibitem[He et~al., 2020]{he2020pathvqa}
He, X., Zhang, Y., Mou, L., Xing, E., and Xie, P. (2020).
\newblock Pathvqa: 30000+ questions for medical visual question answering.
\newblock {\em arXiv preprint arXiv:2003.10286}.

\bibitem[Huang et~al., 2021]{huang2021gloria}
Huang, S.-C., Shen, L., Lungren, M.~P., and Yeung, S. (2021).
\newblock Gloria: A multimodal global-local representation learning framework for label-efficient medical image recognition.
\newblock In {\em Proceedings of the IEEE/CVF International Conference on Computer Vision}, pages 3942--3951.

\bibitem[Huang et~al., 2023]{huang2023visual}
Huang, Z., Bianchi, F., Yuksekgonul, M., Montine, T.~J., and Zou, J. (2023).
\newblock A visual--language foundation model for pathology image analysis using medical twitter.
\newblock {\em Nature medicine}, 29(9):2307--2316.

\bibitem[Ikezogwo et~al., 2024]{ikezogwo2024quilt}
Ikezogwo, W., Seyfioglu, S., Ghezloo, F., Geva, D., Sheikh~Mohammed, F., Anand, P.~K., Krishna, R., and Shapiro, L. (2024).
\newblock Quilt-1m: One million image-text pairs for histopathology.
\newblock {\em Advances in Neural Information Processing Systems}, 36.

\bibitem[Jia et~al., 2021]{jia2021scaling}
Jia, C., Yang, Y., Xia, Y., Chen, Y.-T., Parekh, Z., Pham, H., Le, Q., Sung, Y.-H., Li, Z., and Duerig, T. (2021).
\newblock Scaling up visual and vision-language representation learning with noisy text supervision.
\newblock In {\em International conference on machine learning}, pages 4904--4916. PMLR.

\bibitem[Johnson et~al., 2019]{johnson2019mimic}
Johnson, A.~E., Pollard, T.~J., Berkowitz, S.~J., Greenbaum, N.~R., Lungren, M.~P., Deng, C.-y., Mark, R.~G., and Horng, S. (2019).
\newblock Mimic-cxr, a de-identified publicly available database of chest radiographs with free-text reports.
\newblock {\em Scientific data}, 6(1):317.

\bibitem[Kaplan et~al., 2020]{kaplan2020scaling}
Kaplan, J., McCandlish, S., Henighan, T., Brown, T.~B., Chess, B., Child, R., Gray, S., Radford, A., Wu, J., and Amodei, D. (2020).
\newblock Scaling laws for neural language models.
\newblock {\em arXiv preprint arXiv:2001.08361}.

\bibitem[Kim et~al., 2023]{kim2023concept}
Kim, I., Kim, J., Choi, J., and Kim, H.~J. (2023).
\newblock Concept bottleneck with visual concept filtering for explainable medical image classification.
\newblock In {\em International Conference on Medical Image Computing and Computer-Assisted Intervention}, pages 225--233. Springer.

\bibitem[Kim et~al., 2021]{kim2021vilt}
Kim, W., Son, B., and Kim, I. (2021).
\newblock Vilt: Vision-and-language transformer without convolution or region supervision.
\newblock In {\em International conference on machine learning}, pages 5583--5594. PMLR.

\bibitem[Lau et~al., 2018]{lau2018dataset}
Lau, J.~J., Gayen, S., Ben~Abacha, A., and Demner-Fushman, D. (2018).
\newblock A dataset of clinically generated visual questions and answers about radiology images.
\newblock {\em Scientific data}, 5(1):1--10.

\bibitem[Li et~al., 2021a]{li2021align}
Li, J., Selvaraju, R., Gotmare, A., Joty, S., Xiong, C., and Hoi, S. C.~H. (2021a).
\newblock Align before fuse: Vision and language representation learning with momentum distillation.
\newblock {\em Advances in neural information processing systems}, 34:9694--9705.

\bibitem[Li et~al., 2020]{li2020unimo}
Li, W., Gao, C., Niu, G., Xiao, X., Liu, H., Liu, J., Wu, H., and Wang, H. (2020).
\newblock Unimo: Towards unified-modal understanding and generation via cross-modal contrastive learning.
\newblock {\em arXiv preprint arXiv:2012.15409}.

\bibitem[Li et~al., 2023]{li2023scaling}
Li, Y., Fan, H., Hu, R., Feichtenhofer, C., and He, K. (2023).
\newblock Scaling language-image pre-training via masking.
\newblock In {\em Proceedings of the IEEE/CVF Conference on Computer Vision and Pattern Recognition}, pages 23390--23400.

\bibitem[Li et~al., 2021b]{li2021supervision}
Li, Y., Liang, F., Zhao, L., Cui, Y., Ouyang, W., Shao, J., Yu, F., and Yan, J. (2021b).
\newblock Supervision exists everywhere: A data efficient contrastive language-image pre-training paradigm.
\newblock {\em arXiv preprint arXiv:2110.05208}.

\bibitem[Lin et~al., 2023]{lin2023pmc}
Lin, W., Zhao, Z., Zhang, X., Wu, C., Zhang, Y., Wang, Y., and Xie, W. (2023).
\newblock Pmc-clip: Contrastive language-image pre-training using biomedical documents.
\newblock In {\em International Conference on Medical Image Computing and Computer-Assisted Intervention}, pages 525--536. Springer.

\bibitem[Liu et~al., 2023]{liu2023imitate}
Liu, C., Cheng, S., Shi, M., Shah, A., Bai, W., and Arcucci, R. (2023).
\newblock Imitate: Clinical prior guided hierarchical vision-language pre-training.
\newblock {\em arXiv preprint arXiv:2310.07355}.

\bibitem[Liu et~al., 2024]{liu2024etp}
Liu, C., Wan, Z., Cheng, S., Zhang, M., and Arcucci, R. (2024).
\newblock Etp: Learning transferable ecg representations via ecg-text pre-training.
\newblock In {\em ICASSP 2024-2024 IEEE International Conference on Acoustics, Speech and Signal Processing (ICASSP)}, pages 8230--8234. IEEE.

\bibitem[Loshchilov and Hutter, 2017]{loshchilov2017decoupled}
Loshchilov, I. and Hutter, F. (2017).
\newblock Decoupled weight decay regularization.
\newblock {\em arXiv preprint arXiv:1711.05101}.

\bibitem[Lu et~al., 2024]{lu2024visual}
Lu, M.~Y., Chen, B., Williamson, D.~F., Chen, R.~J., Liang, I., Ding, T., Jaume, G., Odintsov, I., Le, L.~P., Gerber, G., et~al. (2024).
\newblock A visual-language foundation model for computational pathology.
\newblock {\em Nature Medicine}, 30(3):863--874.

\bibitem[M{\"u}ller et~al., 2022]{muller2022joint}
M{\"u}ller, P., Kaissis, G., Zou, C., and Rueckert, D. (2022).
\newblock Joint learning of localized representations from medical images and reports.
\newblock In {\em European Conference on Computer Vision}, pages 685--701. Springer.

\bibitem[Oord et~al., 2018]{oord2018representation}
Oord, A. v.~d., Li, Y., and Vinyals, O. (2018).
\newblock Representation learning with contrastive predictive coding.
\newblock {\em arXiv preprint arXiv:1807.03748}.

\bibitem[Pang et~al., 2023]{pang2023survey}
Pang, T., Li, P., and Zhao, L. (2023).
\newblock A survey on automatic generation of medical imaging reports based on deep learning.
\newblock {\em BioMedical Engineering OnLine}, 22(1):48.

\bibitem[Pelka et~al., 2018]{pelka2018radiology}
Pelka, O., Koitka, S., R{\"u}ckert, J., Nensa, F., and Friedrich, C.~M. (2018).
\newblock Radiology objects in context (roco): a multimodal image dataset.
\newblock In {\em Intravascular Imaging and Computer Assisted Stenting and Large-Scale Annotation of Biomedical Data and Expert Label Synthesis: 7th Joint International Workshop, CVII-STENT 2018 and Third International Workshop, LABELS 2018, Held in Conjunction with MICCAI 2018, Granada, Spain, September 16, 2018, Proceedings 3}, pages 180--189. Springer.

\bibitem[Radford et~al., 2021]{radford2021learning}
Radford, A., Kim, J.~W., Hallacy, C., Ramesh, A., Goh, G., Agarwal, S., Sastry, G., Askell, A., Mishkin, P., Clark, J., et~al. (2021).
\newblock Learning transferable visual models from natural language supervision.
\newblock In {\em International conference on machine learning}, pages 8748--8763. PMLR.

\bibitem[Russakovsky et~al., 2015]{russakovsky2015imagenet}
Russakovsky, O., Deng, J., Su, H., Krause, J., Satheesh, S., Ma, S., Huang, Z., Karpathy, A., Khosla, A., Bernstein, M., et~al. (2015).
\newblock Imagenet large scale visual recognition challenge.
\newblock {\em International journal of computer vision}, 115:211--252.

\bibitem[Seibold et~al., 2022]{seibold2022breaking}
Seibold, C., Rei{\ss}, S., Sarfraz, M.~S., Stiefelhagen, R., and Kleesiek, J. (2022).
\newblock Breaking with fixed set pathology recognition through report-guided contrastive training.
\newblock In {\em International Conference on Medical Image Computing and Computer-Assisted Intervention}, pages 690--700. Springer.

\bibitem[Singh et~al., 2020]{singh2020mmf}
Singh, A., Goswami, V., Natarajan, V., Jiang, Y., Chen, X., Shah, M., Rohrbach, M., Batra, D., and Parikh, D. (2020).
\newblock Mmf: A multimodal framework for vision and language research.
\newblock \url{https://github.com/facebookresearch/mmf}.

\bibitem[Singh et~al., 2022]{singh2022flava}
Singh, A., Hu, R., Goswami, V., Couairon, G., Galuba, W., Rohrbach, M., and Kiela, D. (2022).
\newblock Flava: A foundational language and vision alignment model.
\newblock In {\em Proceedings of the IEEE/CVF Conference on Computer Vision and Pattern Recognition}, pages 15638--15650.

\bibitem[Tiu et~al., 2022]{tiu2022expert}
Tiu, E., Talius, E., Patel, P., Langlotz, C.~P., Ng, A.~Y., and Rajpurkar, P. (2022).
\newblock Expert-level detection of pathologies from unannotated chest x-ray images via self-supervised learning.
\newblock {\em Nature Biomedical Engineering}, 6(12):1399--1406.

\bibitem[Tu et~al., 2024]{tu2024closer}
Tu, W., Deng, W., and Gedeon, T. (2024).
\newblock A closer look at the robustness of contrastive language-image pre-training (clip).
\newblock {\em Advances in Neural Information Processing Systems}, 36.

\bibitem[Wang et~al., 2023]{wang2023one}
Wang, P., Wang, S., Lin, J., Bai, S., Zhou, X., Zhou, J., Wang, X., and Zhou, C. (2023).
\newblock One-peace: Exploring one general representation model toward unlimited modalities.
\newblock {\em arXiv preprint arXiv:2305.11172}.

\bibitem[Wei and Hu, 2024]{wei2024mmpareto}
Wei, Y. and Hu, D. (2024).
\newblock Mmpareto: Boosting multimodal learning with innocent unimodal assistance.
\newblock In {\em International conference on machine learning}. PMLR.

\bibitem[Yao et~al., 2021]{yao2021filip}
Yao, L., Huang, R., Hou, L., Lu, G., Niu, M., Xu, H., Liang, X., Li, Z., Jiang, X., and Xu, C. (2021).
\newblock Filip: Fine-grained interactive language-image pre-training.
\newblock {\em arXiv preprint arXiv:2111.07783}.

\bibitem[Yu et~al., 2022]{yu2022coca}
Yu, J., Wang, Z., Vasudevan, V., Yeung, L., Seyedhosseini, M., and Wu, Y. (2022).
\newblock Coca: Contrastive captioners are image-text foundation models.
\newblock {\em arXiv preprint arXiv:2205.01917}.

\bibitem[Zhai et~al., 2022]{zhai2022lit}
Zhai, X., Wang, X., Mustafa, B., Steiner, A., Keysers, D., Kolesnikov, A., and Beyer, L. (2022).
\newblock Lit: Zero-shot transfer with locked-image text tuning.
\newblock In {\em Proceedings of the IEEE/CVF Conference on Computer Vision and Pattern Recognition}, pages 18123--18133.

\bibitem[Zhan et~al., 2021]{zhan2021product1m}
Zhan, X., Wu, Y., Dong, X., Wei, Y., Lu, M., Zhang, Y., Xu, H., and Liang, X. (2021).
\newblock Product1m: Towards weakly supervised instance-level product retrieval via cross-modal pretraining.
\newblock In {\em Proceedings of the IEEE/CVF International Conference on Computer Vision}, pages 11782--11791.

\bibitem[Zhang et~al., 2024]{zhang2024vision}
Zhang, J., Huang, J., Jin, S., and Lu, S. (2024).
\newblock Vision-language models for vision tasks: A survey.
\newblock {\em IEEE Transactions on Pattern Analysis and Machine Intelligence}.

\bibitem[Zhang et~al., 2023a]{zhang2023biomedclip}
Zhang, S., Xu, Y., Usuyama, N., Xu, H., Bagga, J., Tinn, R., Preston, S., Rao, R., Wei, M., Valluri, N., et~al. (2023a).
\newblock Biomedclip: a multimodal biomedical foundation model pretrained from fifteen million scientific image-text pairs.
\newblock {\em arXiv preprint arXiv:2303.00915}.

\bibitem[Zhang et~al., 2023b]{zhang2023multimodal}
Zhang, X., Yoon, J., Bansal, M., and Yao, H. (2023b).
\newblock Multimodal representation learning by alternating unimodal adaptation.
\newblock {\em arXiv preprint arXiv:2311.10707}.

\bibitem[Zhang et~al., 2023c]{zhang2023text}
Zhang, Y., Gao, J., Zhou, M., Wang, X., Qiao, Y., Zhang, S., and Wang, D. (2023c).
\newblock Text-guided foundation model adaptation for pathological image classification.
\newblock In {\em International Conference on Medical Image Computing and Computer-Assisted Intervention}, pages 272--282. Springer.

\bibitem[Zhao et~al., 2023]{zhao2023clip}
Zhao, Z., Liu, Y., Wu, H., Li, Y., Wang, S., Teng, L., Liu, D., Li, X., Cui, Z., Wang, Q., et~al. (2023).
\newblock Clip in medical imaging: A comprehensive survey.
\newblock {\em arXiv preprint arXiv:2312.07353}.

\end{thebibliography}
